\documentclass[lettersize,journal]{IEEEtran}
\usepackage{amsmath,amsfonts}
\usepackage{algorithmic}
\usepackage{algorithm}
\usepackage{array}
\usepackage[caption=false,font=normalsize,labelfont=sf,textfont=sf]{subfig}
\usepackage{textcomp}
\usepackage{stfloats}
\usepackage{url}
\usepackage{verbatim}
\usepackage{graphicx}
\usepackage{cite}
\usepackage{amsmath}
\usepackage{multirow}
\usepackage{multicol}
\usepackage[table,xcdraw]{xcolor}
\usepackage{amsfonts}
\usepackage{booktabs}
\usepackage{algorithm}

\usepackage{algorithmic}
\usepackage{graphicx}
\usepackage{wrapfig}
\definecolor{myPink}{RGB}{191, 0, 64}
\usepackage[colorlinks=true, allcolors=blue]{hyperref}
\begin{document}

\title{DifferSeg: Towards Diverse Multimodal Binary Segmentation via \\Differential Perception and Frequency Guidance}

\author{Qiangqiang~Zhou, Jiawei~Xu*, Yong~Chen, Dandan~Zhu, Yugen~Yi, and Xiaoqi~Zhao*%
\thanks{Qiangqiang Zhou, Jiawei Xu, Yong Chen, and Yugen Yi are with the School of Artificial Intelligence, Jiangxi Normal University, Nanchang, China
(e-mail: jiawei\_xu@jxnu.edu.cn).}%
\thanks{Dandan Zhu is with the Institute of AI Education, East China Normal University, Shanghai, China
(e-mail: ddzhu@mail.ecnu.edu.cn).}%
\thanks{Xiaoqi Zhao is with the Yale School of Medicine, Yale University, New Haven, CT, USA
(e-mail: xiaoqi.zhao@yale.edu).}%
\thanks{*Corresponding author: Jiawei Xu and Xiaoqi Zhao.}%

\thanks{Manuscript received April 19, 2021; revised August 16, 2021.}
}

\markboth{IEEE TRANSACTIONS ON CIRCUITS AND SYSTEMS FOR VIDEO TECHNOLOGY}%
{Shell \MakeLowercase{\textit{et al.}}: A Sample Article Using IEEEtran.cls for IEEE Journals}


\maketitle

\begin{abstract}
In many binary segmentation tasks, most multimodal methods rely on fixed feature concatenation for cross-modal interaction and straightforward decoder designs dominated by low-frequency semantics.
%
However, they ignore two key challenges: one is the lack of an adaptive mechanism to handle modality discrepancies and complementarity, and the other is the absence of an efficient decoding strategy to balance both high- and low-frequency representations.
In this work, we propose a simple yet general multimodal binary segmentation framework, termed DifferSeg, to address both problems simultaneously.
With the help of the differential perception fusion (DPF) module, DifferSeg employs learnable differential operators to adaptively align multimodal features and enhance their complementarity through residual fusion, effectively mitigating modality mismatch and fusion redundancy.
In addition, we design a frequency-guided decoder (FGD) that builds cross-frequency interactions and multi-path upsampling to maintain consistency between detailed high-frequency structures and semantic low-frequency representations, ensuring fine-grained boundary recovery and noise suppression.
Benefiting from these designs, DifferSeg can be easily generalized to diverse binary segmentation tasks, including both natural and medical modalities. Without bells and whistles, it consistently surpasses 67 state-of-the-art methods across 29 public datasets involving 18 downstream tasks, demonstrating superior generalization and segmentation accuracy. Code and pretrained models will be available at the
\href{https://github.com/jiaweiXu1029/DifferSeg}{\textcolor{myPink}{link}}
.
\end{abstract}


\section{Introduction}
\label{sec:intro}
\IEEEPARstart{B}{inary} segmentation is a fundamental problem in computer vision \cite{Tcsvt0,GateNet}, aiming to distinguish target objects or regions of interest from the background in images or videos. It serves as a core visual perception task and supports a wide range of downstream applications across military, industrial, and medical fields.
Over time, various specialized branches of binary segmentation have been explored across diverse scenarios, such as salient object detection \cite{Tcsvt1,Tcsvt2} and camouflaged object detection \cite{adaptCOD,Tcsvt3} in natural scenes, transparent object segmentation  \cite{eblnet} and defocus blur detection \cite{DD} in industrial scenes, colon polyp \cite{CTNet-M} and skin lesion segmentation \cite{flownet} in different medical imaging.

\begin{figure}[t]
    \centering
     \includegraphics[width=\columnwidth]{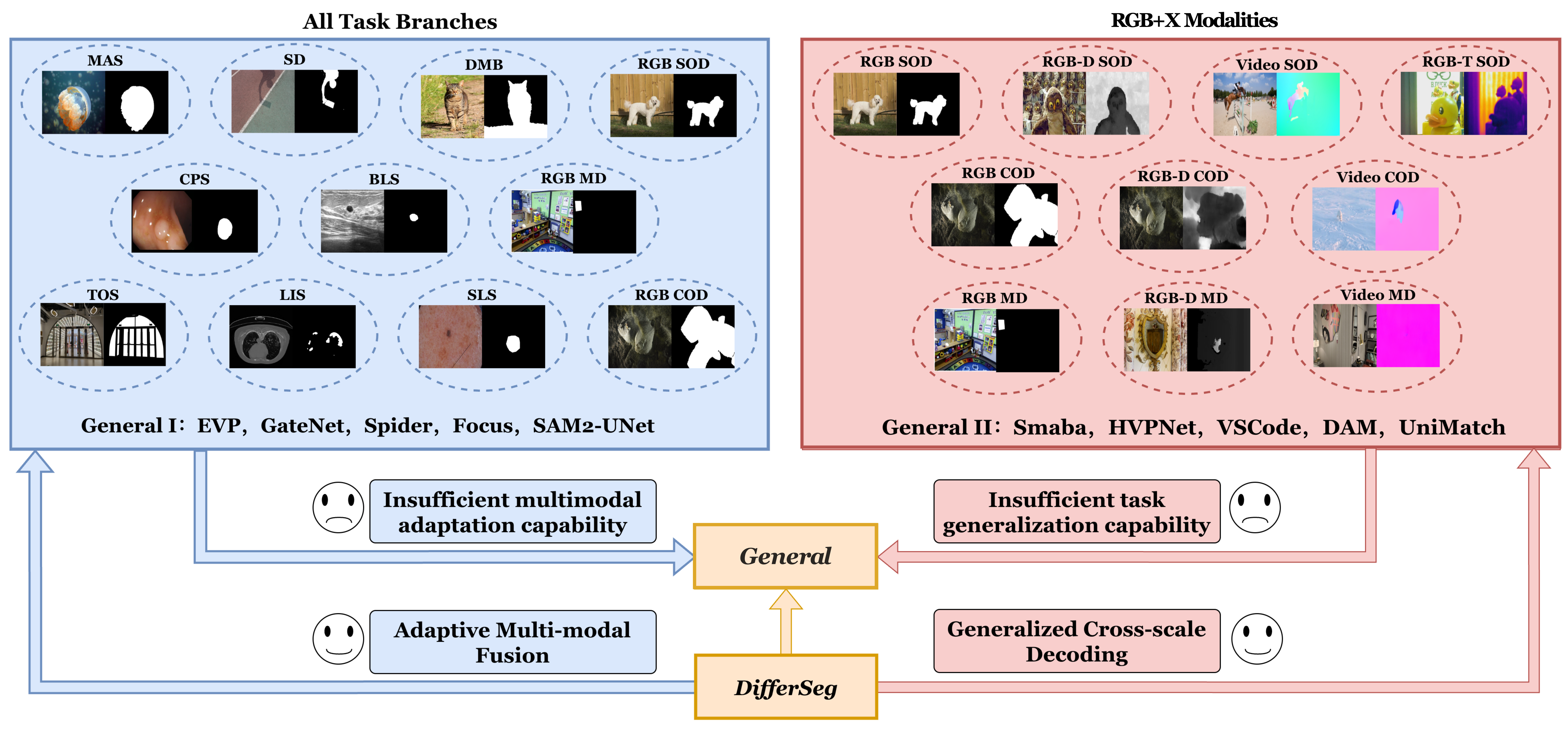}
\caption{Overview of the limitations of existing generalization directions and the motivation of DifferSeg. The left part illustrates the ``All Task Branches'' setting, where \textbf{\textit{General \MakeUppercase{\romannumeral 1}}} methods~\cite{EVP,GateNet,spider,focus,sam2unet,tpseg} operate within a single modality but cannot adapt well to multimodal inputs. The right part illustrates the ``RGB+X Modalities'' setting, where \textbf{\textit{General \MakeUppercase{\romannumeral 2}}} methods~\cite{samba,hvpnet,vscode,DAM,UniMatch} handle RGB+X Modalities but fail to generalize across diverse binary segmentation task branches. DifferSeg overcomes these limitations through adaptive multimodal fusion and generalized cross-scale decoding, enabling general under RGB + X modality setting binary segmentation.}

    \label{fig:image1}
\end{figure}

Although numerous well-established models \cite{TSPSAM,tdfnet,RISNet,sam-cod,hvpnet,TSCVT4,TSCVT7,TSCVT5} have been developed for various task branches of binary segmentation, most are tailored to a single task or modality, which inevitably constrains the overall progress of the field. 
To overcome this fragmentation, recent studies have begun exploring general binary segmentation models \cite{sam2unet,vscode,hvpnet,focus,DAM}, evolving along two main directions: \textbf{\textit{General \MakeUppercase{\romannumeral 1}}} representing general models within a single modality, and \textbf{\textit{General \MakeUppercase{\romannumeral 2}}} representing general models across tasks within a specific task branches.
%
However, neither direction achieves full generality. For instance, FOCUS \cite{focus} (\textbf{\textit{General \MakeUppercase{\romannumeral 1}}}) leverages multi-scale semantics, edge cues, and contrastive distillation to perform well across multiple binary segmentation tasks, yet is restricted to single-modal inputs. Similarly, Samba \cite{samba} (\textbf{\textit{General \MakeUppercase{\romannumeral 2}}}) excels in general salient object detection through its Saliency-Guided Mamba Module and spatial neighborhood scanning, but its design is highly task-specific and does not generalize to other branches. As shown in Fig.~\hyperref[fig:image1]{1}, these limitations highlight the need for a unified framework capable of bridging modalities and tasks to achieve comprehensive generalization across diverse binary segmentation scenarios.
%

Achieving general under RGB + auxiliary modality setting binary segmentation requires addressing two key challenges.
\textit{\textbf{First}}, multimodal inputs such as RGB, depth, thermal, or flow modalities often exhibit severe information inconsistency and incompleteness. RGB images provide rich texture and color cues, while other modalities emphasize complementary structural or semantic information, yet their distributions, noise characteristics, and spatial responses differ significantly. As a result, fixed or hand-crafted fusion strategies tend to overemphasize dominant modalities or suppress subtle yet critical cues, leading to unstable performance when applied across heterogeneous modalities and diverse scenarios. This makes it particularly challenging to design a general fusion mechanism that can adaptively exploit inter-modal complementarity while remaining robust to modality-specific noise and missing information.
\textit{\textbf{Second}}, during cross-scale decoding, frequency-domain aliasing is a pervasive issue. High-frequency components, which are essential for preserving object boundaries and fine structures, are often diluted by dominant low-frequency semantic features propagated from deeper layers. Conversely, low-frequency signals are highly susceptible to noise accumulation during upsampling and feature aggregation. Without explicitly modeling and balancing frequency components, conventional decoders struggle to maintain a consistent representation across scales, resulting in blurred boundaries, false positives, or degraded generalization, especially in complex or low-contrast segmentation scenarios.

To address these challenges, we propose DifferSeg, a general multimodal binary segmentation framework covering diverse tasks across both natural and medical scenarios.
We introduce a differential perception fusion (DPF) module that employs learnable differential operators to adaptively align and enhance inter-modal complementarity during training, enabling RGB to fuse effectively with other modalities. DPF further emphasizes low-frequency semantic information while preserving high-frequency details, achieving robust and general multimodal fusion. In addition, to cope with frequency-domain aliasing in decoding, we design a frequency-guided decoder (FGD) module. FGD explicitly separates high- and low-frequency components at each scale, introduces cross-frequency interactions, and maintains their complementarity through progressive multi-path upsampling with an adaptive frequency balancer, ensuring generalized cross-scale decoding that preserves fine details and suppresses noise.

To comprehensively evaluate the performance of DifferSeg, we conducted rigorous testing on \textbf{29} representative benchmark datasets in the field of binary segmentation. Spanning multiple application domains, these experiments aim to thoroughly validate the model's superior performance and cross-domain generalization. Experimental results demonstrate that DifferSeg significantly outperforms \textbf{67} state-of-the-art (SOTA) methods across key metrics, establishing its leading position in handling complex segmentation tasks.

In summary, our main contributions are as follows:
\begin{itemize}
\item We propose DifferSeg, a general multimodal binary segmentation framework that spans diverse tasks in both natural and medical scenarios, aiming to achieve comprehensive generalization across tasks and modalities.
\item We design a differential perception fusion module that employs learnable differential operators to adaptively align multimodal features, enhancing inter-modal complementarity while preserving both high- and low-frequency information for robust fusion.
\item We introduce a frequency-guided decoder that leverages cross-frequency interactions and multi-path upsampling to preserve high-frequency details while suppressing low-frequency noise, forming a general decoding paradigm. 
\item  Extensive experiments show that DifferSeg consistently outperforms state-of-the-art methods across 18 downstream tasks in natural and medical domains, demonstrating superior accuracy, robustness, and generalization.
\end{itemize}
\section{Related Work}
\label{sec:formatting}
\subsection{Binary Segmentation}
Binary Segmentation~\cite{GateNet} encompasses a wide range of crucial tasks across both natural and medical image domains. In the natural image domain, representative tasks include salient object detection (SOD) \cite{SOMANet}, camouflaged object detection (COD) \cite{sam-cod}, marine animal segmentation (MAS) \cite{mas-sam}, shadow detection (SD) \cite{SARA}, defocus blur detection (DBD) \cite{msu-mamba}, transparent object segmentation (TOS) \cite{rfenet}, and mirror detection (MD) \cite{DAM}. In the medical image domain, foreground segmentation tasks cover colon polyp segmentation (CPS) \cite{CTNet-M}, COVID-19 lung infection detection (CLI) \cite{DECOR-Net}, breast lesion segmentation (BLS) \cite{DBUNet}, and skin lesion segmentation (SLS) \cite{flowsdf}.
\begin{figure*}[t]
    \centering
    \includegraphics[width=1\textwidth]{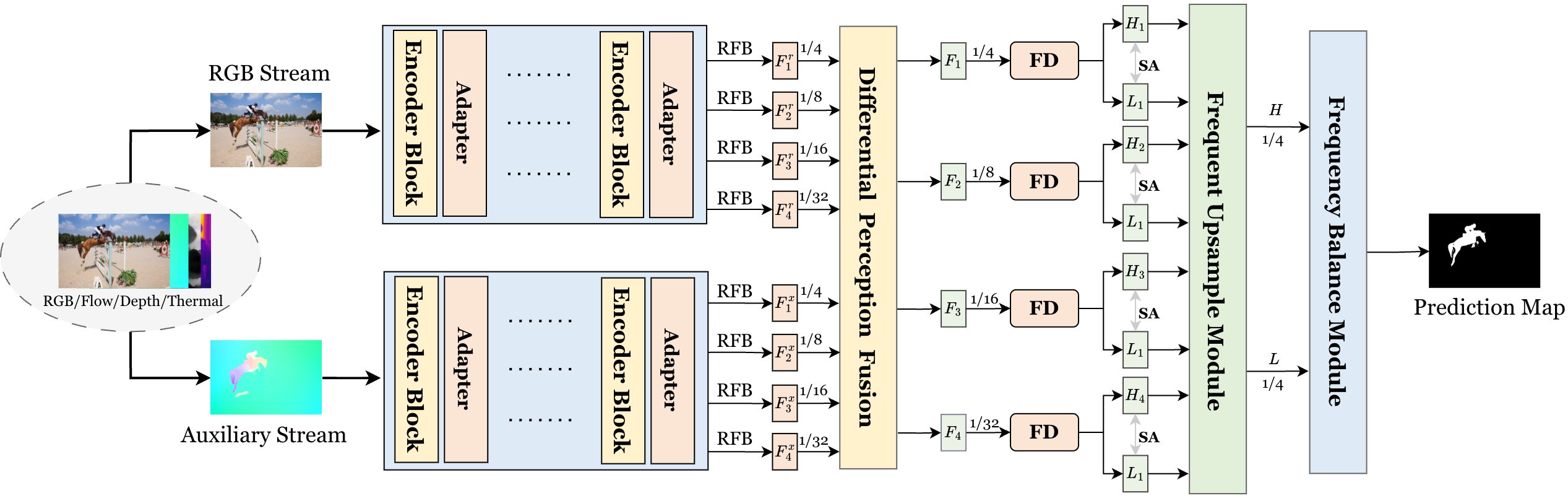}
    \caption{Overall architecture of the proposed DifferSeg for general tasks. The input RGB image and another image are processed by the SAM2 encoder with adapters to extract multi-level features, which are fused by the differential perception fusion (DPF) module to obtain $\{F_i\}_{i=1}^4$. In decoding, frequency decomposition (FD) module decompositions features into high- and low-frequency components, scale attention (SA) module performs cross-frequency interactions, and frequency upsample (FU) module progressively upsamples. The final fused features are refined by the frequency balance (FB) module to produce the prediction map.}
    \label{fig:image2}
\end{figure*}

Each task branch has its specific focus. SOD aims to segment the most visually attractive objects from an image, while COD targets objects that are seamlessly disguised within their environment, such as mimetic animals or body paintings. MAS focuses on separating marine animals from complex ocean backgrounds. SD aims to detect and segment shadow regions from natural scenes. DBD seeks to distinguish between in-focus and out-of-focus regions, usually caused by different focal lengths in imaging devices, while MD identifies the presence and boundaries of mirrors in a scene. TOS, on the other hand, focuses on detecting transparent objects that are often challenging due to their weak contrast with the background. In the medical domain, CPS plays an essential role in assisting early diagnosis of colorectal cancer, while CLI supports the detection of infected lung areas in CT scans for COVID-19 patients. BLS is critical for the precise localization of breast lesions in mammography, and SLS contributes to the early identification of malignant skin diseases such as melanoma.

\subsection{General Binary Segmentation Methods}
\textbf{\textit{General \MakeUppercase{\romannumeral 1}}.} This refers to general models \cite{EVP, GateNet, sam2unet, Controllable-LPMoE,SOMANet, TSCVT8} that address multiple task branches of binary segmentation within a single modality. Some works~\cite{sam2unet, focus} leverage a single architecture to handle various tasks by adapting to different segmentation challenges. EVP \cite{EVP} improves task performance using techniques like adapter-based tuning with minimal additional trainable parameters. GateNet \cite{GateNet} enhances segmentation by employing complex architectures that focus on multi-level information, context, and feature cooperation.

 \textbf{\textit{General \MakeUppercase{\romannumeral 2}}.} This refers to general models within the task branches of binary segmentation, covering tasks across RGB+X modalities. Several studies have proposed methods \cite{UniMatch, depthany, samba, vscode, VST++, vmdnet, hvpnet, FSEL,TSCVT6} in this direction. Some works~\cite{samba, vscode, hvpnet,unisod} achieve full-modal generalization in Salient Object Detection through cross-modal learning, prompt learning, and optimized feature integration to handle different input modalities. Recent works~\cite{SAMDSA, FSEL} achieve generalization in camouflaged object detection by integrating multimodal information using dual-stream adapters and bidirectional knowledge distillation, as well as by jointly exploring frequency and spatial domains through entanglement learning. DAM \cite{DAM} introduces an iterative data engine framework that effectively utilizes unlabeled data to generate pseudo-labels, enhancing their reliability through geometric accuracy scoring and multimodal semantic scoring, achieving generalization in mirror detection.

\subsection{Foundation segmentation model}
Over the past two years, the deep learning community has witnessed the emergence of large-scale pixel-level foundation models~\cite{SAM,SAM2,dino,dinov2}, with the Segment Anything Model (SAM)~\cite{SAM,SAM2} series standing out as a prominent exemplar. To leverage the potent feature extraction capabilities of such models, we employ the SAM 2 image encoder as our backbone. To bridge the domain gap between natural scenes and specialized binary segmentation tasks, we integrate lightweight adapters throughout the encoder stages. This parameter-efficient fine-tuning (PEFT)~\cite{sam2adapter} strategy enables the adaptation of task-specific features with minimal computational overhead while preserving the extensive pre-trained knowledge of SAM 2.

Departing from the generic mask decoders found in vanilla foundation models, we propose a customized architecture tailored for binary segmentation. Specifically, to mitigate information inconsistency and detail loss during multimodal fusion, we introduce the Differential Perception Fusion (DPF) module. By incorporating learnable differential operators (e.g., Sobel and Laplacian), DPF dynamically extracts fine-grained spatial gradients and structural information from RGB and auxiliary modalities (e.g., depth, optical flow). Through spatially adaptive complementarity modeling, DPF effectively synergizes high-frequency details with low-frequency global context, enhancing boundary awareness in challenging environments. Furthermore, to prevent the degradation of high-frequency components and noise contamination during upsampling, we develop the Frequency-Guided Decoder (FGD). The FGD explicitly decouples features into high- and low-frequency components at each scale via a Frequency Decomposition (FD) module. Augmented by scale-attention-driven cross-frequency interaction and a multi-path frequency upsampling strategy, FGD facilitates the recovery of both global structures and sharp boundaries. Finally, a Frequency Balance (FB) module adaptively calibrates the contribution of each frequency component, allowing DifferSeg to achieve robust and high-precision binary segmentation.
\section{Methodology}
\subsection{Overview}
The architecture of DifferSeg, shown in Fig.~\hyperref[fig:image2]{2}, consists of a SAM2~\cite{SAM2} encoder with embedded adapters, a Differential Perception Fusion (DPF) module, and a Frequency-Guided Decoder (FGD). The SAM2 encoder first extracts multi-scale features independently from the RGB image and other modalities (e.g., depth maps, thermal, or optical flow), where lightweight adapters are inserted before each freeze encoder block to accommodate domain differences in downstream tasks. These features are then progressively fused through the DPF, which dynamically enhances modality complementarity while reducing information inconsistencies. The fused representations are subsequently fed into the FGD. At each scale, the frequency decomposition (FD) module explicitly separates high- and low-frequency components, followed by cross-frequency interactions introduced by the scale attention (SA), and progressive refinement via multi-path upsampling. The prediction map is finally produced after adjusting the final scale using the frequency balance (FB) module. 

\subsection{Feature Extraction}
DifferSeg employs the Hiera backbone pretrained by SAM2 \cite{SAM2} to efficiently capture multi-scale features. To better adapt to different task branches of binary segmentation, we insert adapters \cite{sam2adapter} before each multi-scale block of Hiera. Specifically, each adapter consists of a linear layer for downsampling, a GeLU activation function, a linear layer for upsampling, and a final GeLU activation, which maintains representational capacity while improving parameter efficiency. Based on this design, the output multi-scale features are further passed through four receptive field blocks to compress the channel dimension and enhance the representation of these lightweight features.

In multimodal binary segmentation scenarios, the feature extraction process for other modality images (such as depth maps, thermal maps, and optical flow maps) is kept consistent with that of RGB images, ensuring that multimodal information can be effectively integrated within a unified framework. Specifically, given an input image or another image $I \in \mathbb{R}^{3 \times H \times W}$, where $H$ denotes height and $W$ denotes width, our SAM2 encoder produces four hierarchical features $\{F_i^x, F_i^r\} \in \mathbb{R}^{C_i \times \frac{H}{2^{i+1}} \times \frac{W}{2^{i+1}}}$
, where $i \in \{1, 2, 3, 4\}$. For Hiera-L, $C_i \in \{144, 288, 576, 1152\}$. Subsequently, these features are further processed by four receptive field blocks, reducing the channel dimensions to $C_i \in \{64, 128, 256, 512\}$, thereby obtaining more compact and expressive representations.

\begin{figure}[t]
    \centering
    \includegraphics[width=\columnwidth]{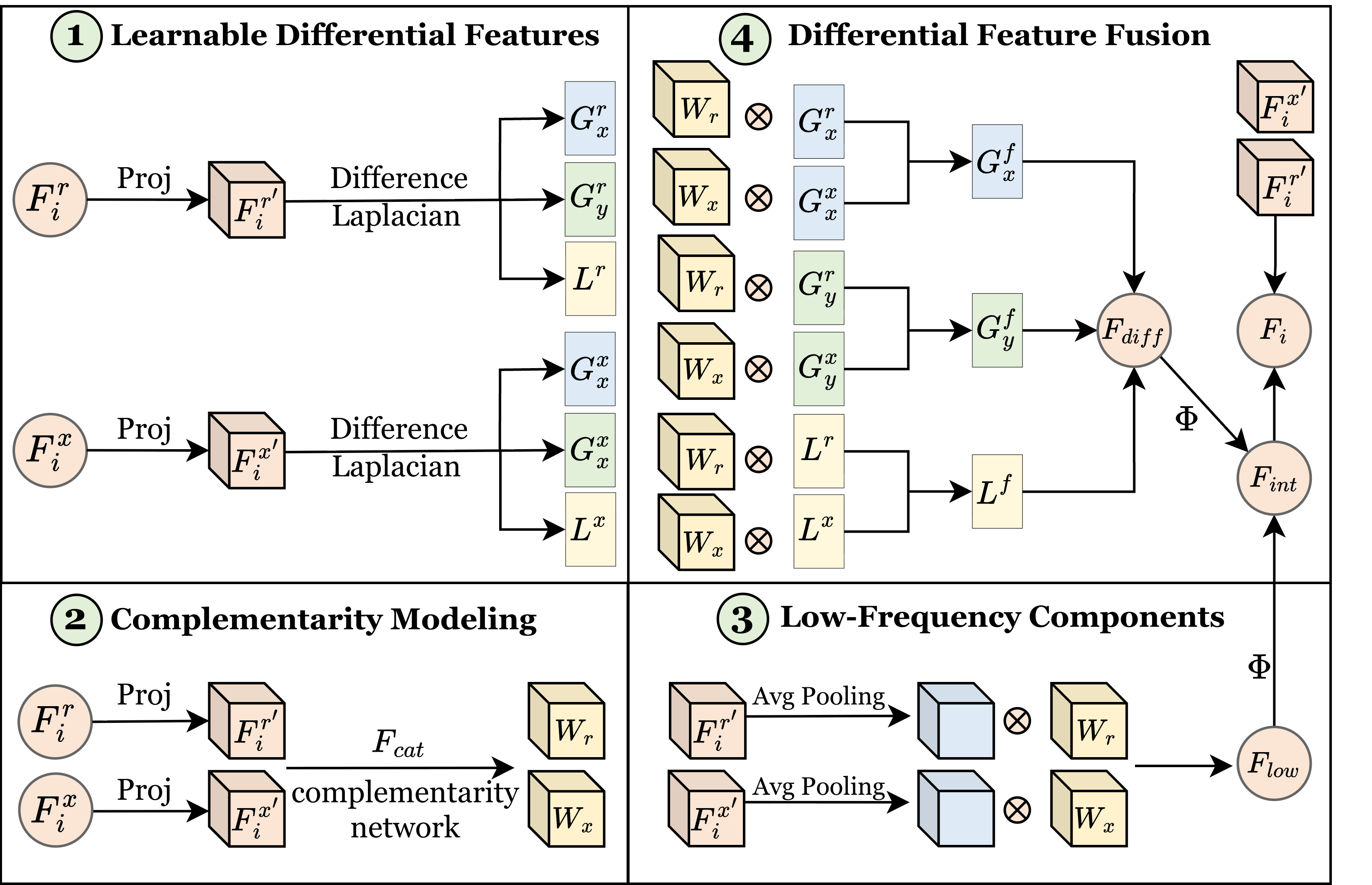}
\caption{Illustration of the differential perception fusion module, which enables effective multimodal fusion through differential feature extraction, adaptive weighted fusion, and residual connections within a unified feature space.}
    \label{fig:image3}
\end{figure}

\begin{figure*}[t]
    \centering
    \includegraphics[width=\textwidth]{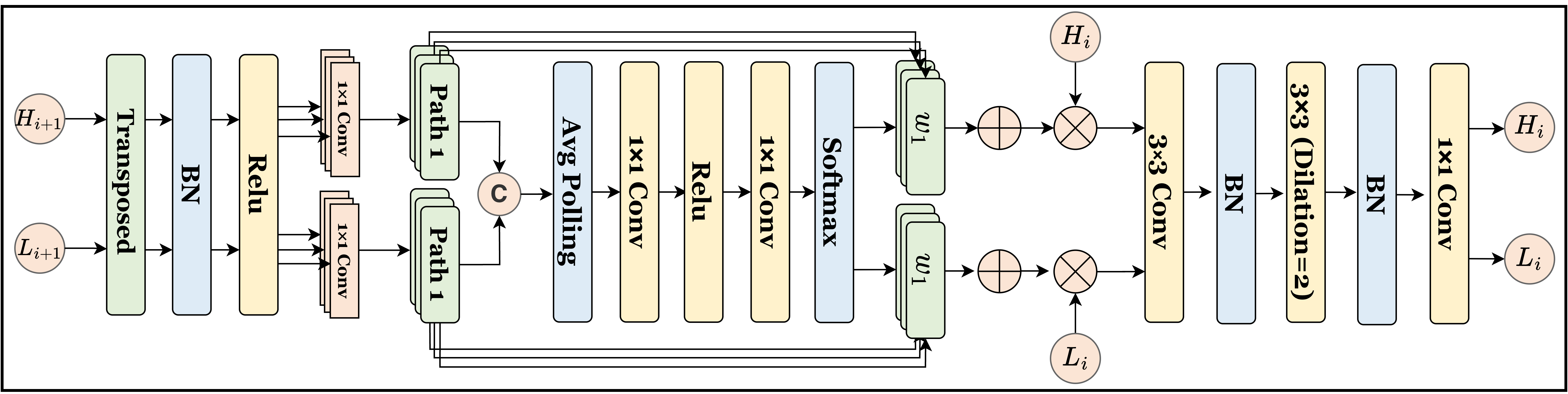}
    \caption{Illustration of the frequency upsample module, which upsamples deep-layer features via transposed convolutions, adaptively fuses them with softmax-normalized weights, and progressively refines them with current-scale features to produce the final low- and high-frequency outputs.}
    \label{fig:image5}
\end{figure*}
\begin{figure}[t]
    \centering
    \includegraphics[width=\columnwidth]{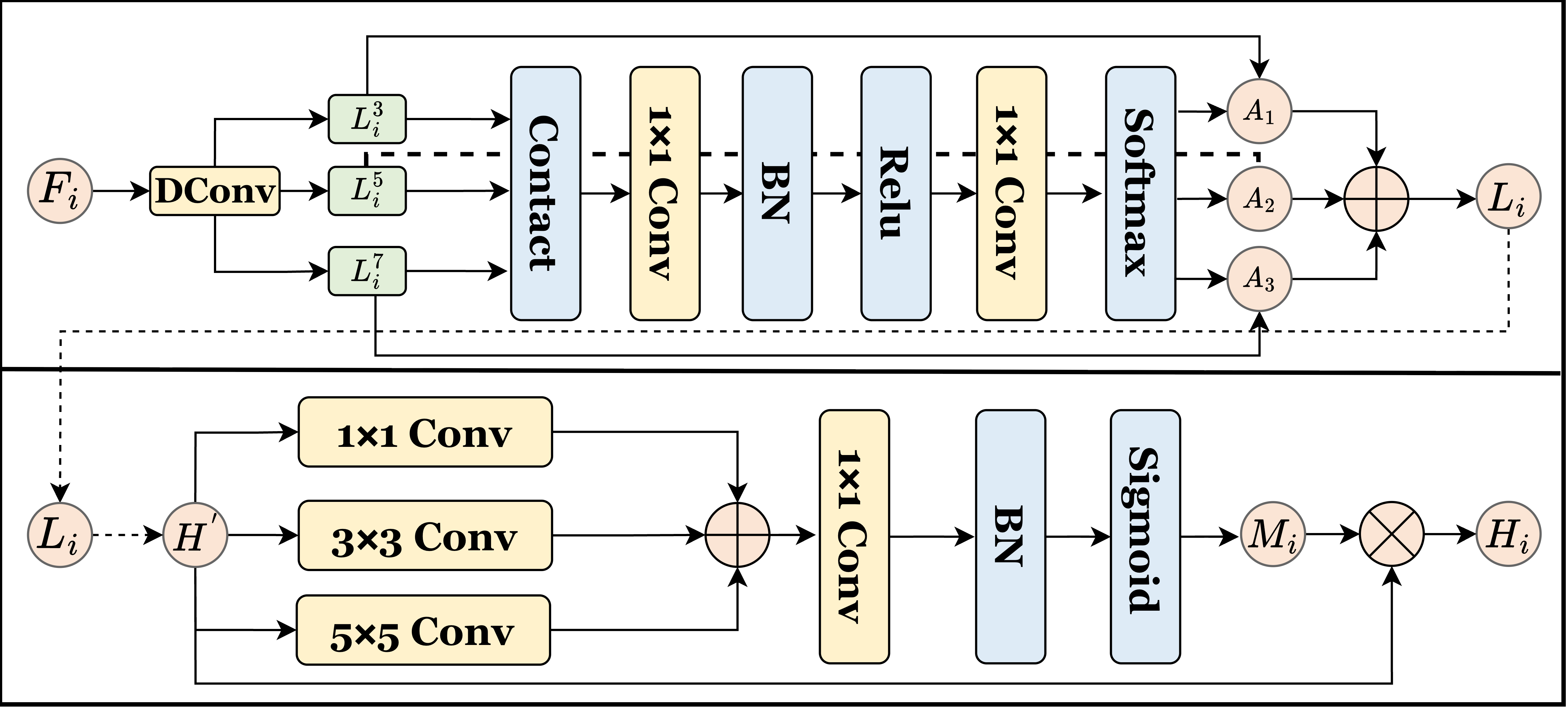}
    \caption{Illustration of the frequency decomposition module, which separates each feature into low- and high-frequency components using multi-scale convolutions and gated enhancement.}
    \label{fig:image4}
\end{figure}
\subsection{Differential Perception Fusion Module}
To address the common issues of multimodal information inconsistency and information loss in the fusion of RGB features and other features, as shown in Fig.~\hyperref[fig:image3]{3}, we propose the innovative differential perception fusion (DPF) module. This module introduces learnable differential operators to dynamically enhance the complementarity between modalities, further optimizing the information fusion process. 
The DPF module not only effectively preserves high-frequency information but also precisely integrates low-frequency information, overcoming the problems of modality mismatch and fusion difficulties commonly encountered in traditional methods. 

\textbf{Learnable Differential Features.}
Given RGB features $\mathbf\{F_i^r\}_{i=1}^4 \in \mathbb{R}^{C \times H \times W}$ and arbitrary modality features $\mathbf\{F_i^x\}_{i=1}^4 \in \mathbb{R}^{C \times H \times W}$, we first align them into a unified feature space through a projection operation:
\begin{equation}
F_{i}^{r'} = \phi_{\text{rgb}}(F_{i}^{r}), \quad F_{i}^{x'} = \phi_{x}(F_{i}^{x}),
\end{equation}
where $\phi$ denotes the projection, which is performed using a $1 \times 1$ convolution operation.

To capture fine-grained spatial variations, we employ learnable differential convolution kernels that are initialized from classical operators: the Sobel-$x$ kernel ${K}_x$, Sobel-$y$ kernel ${K}_y$, and the Laplacian kernel ${K}_{lap}$. The operators in DPF are differential-inspired learnable operators rather than fixed differential filters. We do not impose hard mathematical constraints on these kernels after initialization. Instead, we introduce a differential inductive bias by initializing them with classical Sobel and Laplacian operators. During training, these kernels are adaptively refined while being explicitly used to extract structured responses, including horizontal gradients, vertical gradients, and Laplacian-like structural features, for subsequent complementarity fusion. This design softly encourages the learned kernels to preserve differential-response characteristics while allowing task-specific adaptation. Specifically, for the RGB feature $F_{i}^{r'}$, the differential features are computed as:
\begin{equation}
{G}_x^r = {K}_x * {F_i^{r'}}, \quad {G}_y^r = {K}_y * {F_i^{r'}}, \quad {L}^r = {K}_{lap} * {F_i^{r'}},
\end{equation}
where ${G}_x^r$ and ${G}y^r$ capture directional edge information, while ${L}^r$ extracts second-order variations such as corners and complex structures. Analogously, we apply the same operations to the other modality feature $F_{i}^{x'}$ to obtain:
\begin{equation}
{G}_x^x = {K}_x * {F_{i}^{x'}}, \quad {G}_y^x = {K}_y * {F_{i}^{x'}}, \quad {L}^x = {K}_{lap} * {F_{i}^{x'}}.
\end{equation}

\textbf{Complementarity Modeling.}
Having extracted differential features from both modalities, we now model their complementarity to achieve effective fusion. The projected features are first concatenated as $F_{\text{cat}} = [F_{i}^{r'}, F_{i}^{x'}] \in \mathbb{R}^{2C \times H \times W}$ and fed into a complementarity network $\Psi(\cdot)$, which comprises convolution and batch normalization layers. To dynamically balance the contributions of each modality, we apply softmax normalization to generate spatially-adaptive weights $W_r$ and $W_x$:
\begin{equation}
W = \Psi(F_{\text{cat}}), \quad W = [W_r, W_x],
\end{equation}
with the constraint $W_r(x, y) + W_x(x, y) = 1, \quad \forall(x, y)$.

Leveraging these learned weights, we perform weighted fusion of the differential features from both modalities. The fused differential features $F_{\text{diff}} = [G_x^f, G_y^f, L^f]$ are computed as:
\begin{equation}
G_x^f = W_r \odot G_x^r + W_x \odot G_x^x,
\end{equation}
\begin{equation}
G_y^f = W_r \odot G_y^r + W_x \odot G_y^x,
\end{equation}
\begin{equation}
L^f = W_r \odot L^r + W_x \odot L^x,
\end{equation}
where $\odot$ denotes the element-wise multiplication. This weighted combination enables adaptive integration of directional gradients and structural patterns across modalities.

\textbf{Low-Frequency Components.}
While the differential features capture high-frequency details, it is equally important to preserve low-frequency information for global consistency. To this end, we extract low-frequency components via average pooling $\mathcal{P}(\cdot)$ and combine them using the same complementarity weights:
\begin{equation}
{F}_{low} = {W}_{r} \odot \mathcal{P}({F_i}^{r'}) + {W}_{x} \odot \mathcal{P}({F_i}^{x'}),
\end{equation}
where ${W}$ denotes the spatially averaged weights. This operation stabilizes the fusion process by suppressing noise through spatially-weighted averaging, thereby ensuring robust global feature representation.

\textbf{Differential Feature Fusion.} 
In the fusion process, the fused representation ${F}_{int}$ is obtained by integrating the differential features ${F}_{diff}$ and the low-frequency information ${F}_{low}$ as:
\begin{equation}
{F}_{int} = \Phi\left([{F}_{diff}, {F}_{low}]\right),
\end{equation}
where $\Phi$ denotes the integration operation. Then, through a residual connection, the fused features are combined with the original projected features to obtain the final output: 
\begin{equation}
{F}_{i} = {F}_{int} + {F}_r' + {F}_x'.
\end{equation}
 This residual fusion not only preserves the unique modal information from the RGB and other modality features but also effectively combines high-frequency and low-frequency components, thereby enhancing the overall performance of the fusion result.

\subsection{Frequency-Guided Decoder Module}
To prevent high-frequency details from being suppressed and low-frequency components from being contaminated by noise, we introduce the frequency-guided decoder (FGD) module. It decomposes features into low-frequency and high-frequency components at each scale and preserves their complementarity during upsampling, recovering global structures and fine boundaries while reducing frequency aliasing.

\textbf{Frequency Decomposition.} As illustrated in Fig.~\hyperref[fig:image4]{5}, given the four-level input features $\mathbf\{F_i\}_{i=1}^4 \in \mathbb{R}^{C \times H \times W}$, the FGD module first utilizes the frequency decomposition (FD) Module to decompose each feature into low-frequency and high-frequency components, denoted as $L_i$ and $H_i$, respectively. Specifically, for each input feature $F_i \in \mathbb{R}^{B \times C_i \times H_i \times W_i}$, depthwise separable convolutions with kernel sizes ${5, 7, 9}$ are applied. These larger kernels act as low-pass filters by covering a larger receptive field, allowing them to capture smoother, broader patterns in the data and effectively extract low-frequency components. The outputs of these convolutions, denoted as $L_i^{(5)}, L_i^{(7)}, L_i^{(9)}$, represent the low-frequency responses of the input features at different scales. These low-pass responses are then fused using softmax-normalized weights $A_i$ as:
\begin{equation}
L_i = \sum_{k \in {5, 7, 9}} A_i^{(k)} \odot L_i^{(k)},
\end{equation}
The high-frequency component is obtained by residual subtraction:
\begin{equation}
H_i^{\text{raw}} = F_i - L_i,
\end{equation}
and is further enhanced through multi-branch convolution with a gating mask $M_i$, resulting in: 
\begin{equation}
H_i = M_i \odot H_i^{\text{raw}}.
\end{equation}

\textbf{Scale Attention.} After frequency decomposition, the scale attention module computes channel-wise weights for both low- and high-frequency components, and introduces a lightweight cross-frequency interaction.
The interaction feature, denoted as $\Phi_i$, is obtained by a depthwise-style convolution over the element-wise product of low- and high-frequency features:
\begin{equation}
\Phi_i = \mathrm{Conv}(L_i \odot H_i),
\end{equation}
 and the enhanced features are given by:
 \begin{equation}
L_i = L_i \odot s_i^{\text{low}} + \lambda_i \Phi_i, \quad
H_i = H_i \odot s_i^{\text{high}} + \lambda_i \Phi_i,
\end{equation}
where $\lambda_i$ is an adaptive parameter, and $s_i^{\text{low}}$, $s_i^{\text{high}}$ are channel attention weights produced by global average pooling followed by two $1 \times 1$ conv.

\textbf{Frequency Upsampling.} The frequency upsample module performs progressive decoding using a multi-path strategy. As illustrated in Fig.~\hyperref[fig:image5]{4}, given the current low-frequency and high-frequency features $\mathbf\{H_{i+1}\}_{i=1}^3 \in \mathbb{R}^{C \times H \times W}$ and $\mathbf\{L_{i+1}\}_{i=1}^3 \in \mathbb{R}^{C \times H \times W}$, they are processed through three upsampling paths (Path 1, 2, 3), with each path using transposed convolution and transformations dependent on convolution kernels of sizes {1, 3, 5}. The concatenated output is adaptively fused using softmax-normalized global weights, where the weighted sum of low- and high-frequency features is represented as:
\begin{equation}
H = \sum_{j=1}^3 w_j H^{(j)}, \quad
L = \sum_{j=1}^3 w_j L^{(j)}.
\end{equation}

The fused results are then combined with the features from the current scale $({L}_{i}, {H}_{i})$ and refined through convolution, batch normalization, ReLU, and dilated convolutions for detail recovery, producing $({L}_{i}, {H}_{i})$ at the matched resolution. This process is repeated at subsequent levels until the final outputs $L$ (1/4) and $H$ (1/4) are obtained.

\textbf{Frequency Balance.}
To balance the contributions of the upsampled low- and high-frequency features $L$ and $H$, we introduce a frequency balance module that predicts their fusion weights from global context:
\begin{equation}
p = \text{softmax}(\text{GAP}(L, H)),
\end{equation}
where GAP denotes global average
pooling. The weighted fusion is computed as:
\begin{equation}
F_{\text{fuse}} = {Conv}([{p}_1 L | {p}_2 H]),
\end{equation}
where ${p}1$ and ${p}2$ represent the predicted weights for the low and high-frequency components, respectively. followed by a residual refinement:
\begin{equation}
F_{final} = F_{fuse} + L + H,
\end{equation}
where $F_{final}$ represents the prediction map. This formulation enables adaptive emphasis of low- and high-frequency components and yields a more balanced final prediction.

\subsection{Loss Function}
Following previous works~\cite{EFANet,sam2unet}, we employ a hybrid structure loss $L_{str}$ consisting of weighted Intersection-over-Union (wIoU)~\cite{IOU} and weighted binary cross-entropy (wBCE)~\cite{CELoss} losses. To adaptively focus on hard pixels near boundaries, we calculate a weight map $W$ based on the ground truth $G$:
\begin{equation}
    W = 1 + 5 \cdot | \text{AvgPool}(G) - G |,
\end{equation}
where $\text{AvgPool}(\cdot)$ denotes a $31 \times 31$ average pooling. The $L_{str}$ effectively mitigates foreground–background imbalance and enhances boundary accuracy by combining region-level overlap with pixel-level supervision.

Furthermore, we apply deep supervision to all four segmentation outputs $S_i$. The final training loss $L_{total}$ is the sum of individual losses across all scales:
\begin{equation}
    L_{total} = \sum_{i=0}^{3} L_{str}(S_i, G).
\end{equation}
\section{Experiment}
\label{experiment}
\subsection{Experimental Setup}
\begin{table*}[t]
    \centering 
    \caption{Quantitative comparison of our DifferSeg against other state-of-the-art approaches across 14 downstream natural image tasks. These tasks include salient object detection (SOD), camouflaged object detection (COD), marine animal segmentation (MAS), shadow detection (SD), defocus blur detection (DBD), transparent object segmentation (TOS), and mirror detection (MD). \textbf{\textit{\MakeUppercase{\romannumeral 1}}} denotes \textbf{\textit{General \MakeUppercase{\romannumeral 1}}} methods, \textbf{\textit{\MakeUppercase{\romannumeral 2}}} denotes \textbf{\textit{General \MakeUppercase{\romannumeral 2}}} methods, and ``–'' indicates specialized methods. The top two results are highlighted in $\textbf{bold}$ and $\underline{\text{underline}}$.}
    \setlength{\tabcolsep}{1mm} 
\resizebox{\linewidth}{!}{
\begin{tabular}{r|c|cccc|cccc|cccc|r|c|cccc|cccc|cccc}
\specialrule{1.5pt}{0pt}{0pt}
\multicolumn{14}{c|}{\textbf{RGB SOD}} & \multicolumn{14}{c}{\textbf{RGB-D SOD}} \\
\midrule
\multirow{2}{*}{Methods}
&\multirow{2}{*}{Type}  
& \multicolumn{4}{c|}{DUTS\cite{DUTS}} 
& \multicolumn{4}{c|}{DUT-O\cite{DUT-O}}  
& \multicolumn{4}{c|}{HKU-IS\cite{HKU-IS}} &\multirow{2}{*}{Methods}  
&\multirow{2}{*}{Type}  
& \multicolumn{4}{c|}{STERE\cite{STERE}} 
& \multicolumn{4}{c|}{NLPR\cite{NLPR}}  
& \multicolumn{4}{c}{DUTLF\cite{DUT-Depth}}\\
&
& $S_m \uparrow$ & $F_m \uparrow$ & $E_m \uparrow$& $M \downarrow$
& $S_m \uparrow$ & $F_m \uparrow$ & $E_m \uparrow$& $M \downarrow$
& $S_m \uparrow$ & $F_m \uparrow$ & $E_m \uparrow$& $M \downarrow$
&
&
& $S_m \uparrow$ & $F_m \uparrow$ & $E_m \uparrow$& $M \downarrow$
& $S_m \uparrow$ & $F_m \uparrow$ & $E_m \uparrow$& $M \downarrow$
& $S_m \uparrow$ & $F_m \uparrow$ & $E_m \uparrow$& $M \downarrow$
\\ 
\hline 
EVP(CVPR'23)~\cite{EVP} &\textbf{\MakeUppercase{\romannumeral 1}}& .917 & .910 & .956 &.027 & .874 & .823 & .905&.045 & .935 & .945 & .971 &.024 &
VST++(TPAMI'24)~\cite{VST++}
&\textbf{\MakeUppercase{\romannumeral 2}} & .921 & .916 & .954 &.034 & .935 & .925 & .964 &.021 & .945 & .950 & .969 &.024 \\

GateNet(IJCV'24)~\cite{GateNet} &\textbf{\MakeUppercase{\romannumeral 1}}& .906 & .911 & .931 & .030 & .847 & .824 & .882 & .050 & .931 & 948 & .965 & .025
&GateNet(IJCV'24)~\cite{GateNet}
&\textbf{\MakeUppercase{\romannumeral 1}}& .921 & .929 & .952 &.033 & .938 & .937 & .966 &.020 & .943 & .957& .969  &.025\\

VSCode(CVPR'24)~\cite{vscode} &\textbf{\MakeUppercase{\romannumeral 2}}& .926 & .922 & .960 &.024& .877 & .840 & .912&.043 & .940 & .951 & .974 &.021 
&VSCode(CVPR'24)~\cite{vscode} &\textbf{\MakeUppercase{\romannumeral 2}}& .928 & .926 & .957 &.030 & .938 & .930 & .966 &.020 & .956 & .964 & .976 &.017 \\

SAM2UNet (ICCVW'25)~\cite{sam2unet} &\textbf{\MakeUppercase{\romannumeral 1}}& .922 & .910 & .957 & .023 & .884 & .852 & .912 & .039 & .937 & .948 & .971 & .021&SAM2UNet (ICCVW'25)~\cite{sam2unet} 
&\textbf{\MakeUppercase{\romannumeral 1}}& .924 & .921 & .947 &.031 & .874 & .823 & .905&.045 & .935 & .945 & .971 &.024\\

UniSOD (TCSVT'25)~\cite{unisod} &\textbf{\MakeUppercase{\romannumeral 2}}& .925 & .906 & .938 & .021 & .876 & .822 & .910 & .037 & .940 & .939 & .969 & .018&UniSOD (TCSVT'25)~\cite{unisod} 
&\textbf{\MakeUppercase{\romannumeral 2}}& .924 & .908 & .937 &.028 & .931 & .909 & .964&.019 & .952 & .952 & .975 &.017\\

Samba (CVPR'25)~\cite{samba} &\textbf{\MakeUppercase{\romannumeral 2}}&  \underline{.932} &  \underline{.930} & \textbf{.966} & \underline{.020}&  \underline{.889} &  \underline{.859} &  \underline{.922}& \underline{.037} &  \underline{.945} &  \underline{.956} &  \underline{.978} & \underline{.018}
&Samba(CVPR'25)~\cite{samba}&\textbf{\MakeUppercase{\romannumeral 2}}
&  \underline{.935} &  \underline{.933} & \underline{.963} & \underline{.026} & \textbf{.947} &  \underline{.941} & \textbf{.976} &\textbf{.014} &  .956 &  .964 &  .976 & .017 \\

VSCode-V2 (TPAMI'26)~\cite{vscode2} &\textbf{\MakeUppercase{\romannumeral 2}}&  .930 &  \underline{.930} & .964 & \underline{.020}&  .885 & .847 &  .917& .038 &  .939 &  .948 &  .971 &.019
&VSCode-V2 (TPAMI'26)~\cite{vscode2}&\textbf{\MakeUppercase{\romannumeral 2}}
&  .930 &  .928 & .956 & .028 & \underline{.946} &  .940 & .973 &.016 &  \textbf{.962} &  \textbf{.969} & \textbf{.980} & \underline{.015} \\
\rowcolor[rgb]{0.918,0.965,0.973}
Ours& &\textbf{.936} &\textbf{.935}& \underline{.965}&\textbf{.019}& \textbf{.895}& \textbf{.864} &\textbf{.926}&\textbf{.033} &\textbf{.948} & \textbf{.959} & \textbf{.980}&\textbf{.017}
&Ours&& \textbf{.939} &\textbf{.941} & \textbf{.966} &\textbf{.023} &  .945 & \textbf{.942} &  \underline{.975}& \underline{.015} &\underline{.958} & \textbf{.969} &\underline{.977} &\textbf{.014}\\
\rowcolor[rgb]{0.918,0.965,0.973}
Average Improvement& &.014 &.018&.012&.004&.019&.026&.018&.008&.010&.012&.009&.003&Average Improvement&&.013&.033&.014&.007&.016&.027&.016&.007&.009&.012&.004&.005 \\
\hline
\end{tabular}
\label{tab:I}
}
\vspace{0.8em} 

\resizebox{\linewidth}{!}{
\begin{tabular}{r|c|cccc|cccc|cccc|r|c|cccc|cccc|cccc}
\specialrule{1.5pt}{0pt}{0pt}
\multicolumn{14}{c|}{\textbf{RGB-T SOD}} & \multicolumn{14}{c}{\textbf{VSOD}} \\
\midrule
\multirow{2}{*}{Methods}
 &\multirow{2}{*}{Type}  
& \multicolumn{4}{c|}{VT821\cite{VT821}} 
& \multicolumn{4}{c|}{VT1000\cite{VT1000}}  
& \multicolumn{4}{c|}{VT5000\cite{VT5000}} &\multirow{2}{*}{Methods}  
&\multirow{2}{*}{Type}  
& \multicolumn{4}{c|}{FBMS\cite{FBMS}} 
& \multicolumn{4}{c|}{SegV2\cite{DAVIS}}  
& \multicolumn{4}{c}{DAVSOD-easy\cite{DAVSOD}}\\

&
& $S_m \uparrow$ & $F_m \uparrow$ & $E_m \uparrow$& $M \downarrow$
& $S_m \uparrow$ & $F_m \uparrow$ & $E_m \uparrow$& $M \downarrow$
& $S_m \uparrow$ & $F_m \uparrow$ & $E_m \uparrow$& $M \downarrow$
&
&
& $S_m \uparrow$ & $F_m \uparrow$ & $E_m \uparrow$& $M \downarrow$
& $S_m \uparrow$ & $F_m \uparrow$ & $E_m \uparrow$& $M \downarrow$
& $S_m \uparrow$ & $F_m \uparrow$ & $E_m \uparrow$& $M \downarrow$
\\ 
\hline 
VST++(TPAMI'24)~\cite{VST++}&\textbf{\MakeUppercase{\romannumeral 2}}& .897 & .868 & .925 &.033 & .940 & .931 & .971 &.020 & .901 & .861 & .936 &.034 &CSTFormer (TNNLS'23)~\cite{CoSTFormer}&-& .869 & .861 & .913 &.045 & .874 & .813 & .943 &.017 & .779 & .667 & .819 &.059 
\\
VSCode(CVPR'24)~\cite{vscode} &\textbf{\MakeUppercase{\romannumeral 2}}& .921 & .906 & \underline{.951} &\underline{.021}& .949 & \underline{.944}&  .981 &\textbf{.017} &  .925 & .892 & .954 & .026 &EVP(CVPR'23)~\cite{EVP} &\textbf{\MakeUppercase{\romannumeral 1}}& .858 & .876 & .924 &.046 & .910 & .901 & .941 &.015 & .764 & .673 & .787 &.071 \\

DIM-SOD(AAAI'25)~\cite{dimsod} &\textbf{\MakeUppercase{\romannumeral 2}}&  .923 & \underline{.917} & .949 & .025 & \textbf{.953} & .935 &  {.955} & .020 & .921 & .898 &  .959& .029
&VSCode(CVPR'24)~\cite{vscode} &\textbf{\MakeUppercase{\romannumeral 2}}& .905 & .902 & .939 &.029 & \textbf{.946} & .937 & .984 &.008 & .800 & .710 & .835 &.052\\

SAM2UNet(ICCVW'25)~\cite{sam2unet} &\textbf{\MakeUppercase{\romannumeral 1}}& .904 & 885 & .934 & .029 & .933 & .921 & .956 & .025 & .920 & 897 & .951 & .031
&SAM2UNet(ICCVW'25)~\cite{sam2unet} &\textbf{\MakeUppercase{\romannumeral 1}}& .873 & .904 & .944 &.036 & .921 & .901 & .962 &.012 & .786 & .688 & .798 &.063\\

PCNet(TIP'25)~\cite{PCNet} &\textbf{\MakeUppercase{\romannumeral 2}}& .915 & .879 & .941 &.029& .943 & .924 & .958 &.021 & .920 & .899 & .956 &.030
&Samba(CVPR'25)~\cite{samba} &\textbf{\MakeUppercase{\romannumeral 2}}& \underline{ .925} & \underline{ .922} &  \underline{.954} & \underline{.022} & .943 &  \underline{.938} &\textbf{.987} &\textbf{.006} &  \underline{.813} &  \underline{.734} &  \underline{.856} &\textbf{.043} \\

VSCode-V2 (TPAMI'26)~\cite{vscode2}&\textbf{\MakeUppercase{\romannumeral 2}}& \textbf{.927} & \textbf{.918} & \textbf{.953} &\textbf{.020}& \textbf{.951} & \textbf{.947} & \textbf{.982} &\textbf{.017} & \underline{.928} &  \textbf{.906} & \underline{.960} &\underline{.023}
&VSCode-V2 (TPAMI'26)~\cite{vscode2} &\textbf{\MakeUppercase{\romannumeral 2}}&  .909 & .913 & .950 & .024 & .944 &  .932 &.979 &\underline{.007} & .806 &  .713 &  .837 &.048\\

\rowcolor[rgb]{0.918,0.965,0.973}
 Ours && \underline{.925} & .908 & \underline{.951} & \underline{.022} &  .950 &  .941 & \textbf{.982}&\textbf{.017}& \textbf{.930} &\underline{.905} & \textbf{.964} &\textbf{.022}
 &Ours && \textbf{.930} & \textbf{.928} & \textbf{.957} &\textbf{.021} &  \underline{.945} & \textbf{.939} &  \underline{.984} & \underline{.007} & \textbf{.815} & \textbf{.737} &\textbf{.858} & \underline{.044}\\

\rowcolor[rgb]{0.918,0.965,0.973}
Average Improvement& &.011 &.013&.011&.004&.006&.008&.015&.003&.011&.013&.013&.006&Average Improvement&&.041&.032&.020&.012&.022&.036&.018&.003&.024&.040&.036&.012 \\
\hline
\end{tabular}
\label{tab:I}
}
\vspace{0.8em} 

\resizebox{\linewidth}{!}{
\begin{tabular}{r|c|ccc|ccc|ccc|r|c|ccc|ccc|ccc|r|c|ccc|ccc}
\specialrule{1.5pt}{0pt}{0pt}
\multicolumn{11}{c|}{\textbf{RGB COD}} & \multicolumn{11}{c|}{\textbf{RGB-D COD}}
& \multicolumn{8}{c}{\textbf{VCOD}}
\\
\midrule
\multirow{2}{*}{\textbf{Methods}}
&\multirow{2}{*}{Type}
& \multicolumn{3}{c|}{CAMO\cite{CAMO}} 
& \multicolumn{3}{c|}{COD10K\cite{COD10K}} 
& \multicolumn{3}{c|}{NC4K\cite{NC4K}}
& \multirow{2}{*}{\textbf{Methods}}
& \multirow{2}{*}{\textbf{Type}}
& \multicolumn{3}{c|}{CAMO\cite{CAMO}} 
& \multicolumn{3}{c|}{COD10K\cite{COD10K}} 
& \multicolumn{3}{c|}{NC4K\cite{NC4K}}
&\multirow{2}{*}{\textbf{Methods}}
&\multirow{2}{*}{\textbf{Type}}
& \multicolumn{3}{c|}{MoCA\cite{MoCA-Mask} }
& \multicolumn{3}{c}{CAD\cite{CAD}}   \\
&
& $S_m\uparrow$ & $E_m \uparrow$& $M \downarrow$
& $S_m\uparrow$  & $E_m \uparrow$& $M \downarrow$
& $S_m\uparrow$ & $E_m \uparrow$& $M \downarrow$
&
&
& $S_m\uparrow$ & $E_m \uparrow$& $M \downarrow$
& $S_m\uparrow$ & $E_m \uparrow$& $M \downarrow$
& $S_m\uparrow$ & $E_m \uparrow$& $M \downarrow$
&
&
& $S_m\uparrow$ & $E_m \uparrow$& $M \downarrow$
& $S_m\uparrow$ & $E_m \uparrow$& $M \downarrow$\\ \hline 
 GateNet(IJCV'24)~\cite{GateNet} &\textbf{\MakeUppercase{\romannumeral 1}}& .829 & .888 &.069& .846 & .901&.028& .869 & .918&.040
 &PopNet(ICCV'23)~\cite{PopNet} &-& .769 & .657&.071  & .840  & .666&.031 & .840 & .726&.043
 &IMEX(TMM'24)~\cite{IMEX}&-& \textbf{.695} &.654& .030 & .633 & 683 & .033 \\

FSEL(ECCV'24)~\cite{FSEL}&- & .833& .893 &.067& \underline{.898} & .903&.031 &  \underline{.914} & .855&.042
&DAINet(MS'24)~\cite{DAINet}&- & .868 & .770 &.079&  .896 & .734&.029&  \underline{.905} & .812&.050
&FSEL(ECCV'24)~\cite{FSEL}&\textbf{\MakeUppercase{\romannumeral 2}}& .596 & .645 & .053& .649 & .702 & .053\\

VSCode(CVPR'24)~\cite{vscode} &\textbf{\MakeUppercase{\romannumeral 2}}& .873  &  .938 &.047& .869  &.942&.024 & .891 & .944&.032
&RISNet(CVPR'24)~\cite{RISNet}&-& .853 & .911&.053  & .832 & .669 &.040& .845 & .741 &.063
&TSPSAM(CVPR'24) \cite{TSPSAM} &-& .689 &  .737 & \textbf{.008} & 704 & .734& .028\\

Adapt-COD(IJCV'25)~\cite{adaptCOD} &-&  \underline{.886}  & .936 & \underline{.043} & .892 &  .943& .021 & .906 &  \underline{.944} & \underline{.029} &VSCode(CVPR'24)~\cite{vscode} &\textbf{\MakeUppercase{\romannumeral 2}}&  .882 & .942 & .044& .874 & .945&.023 & .893&  \underline{.946}&.030 
&SAM-PM(CVPR'24)~\cite{sam-pm} &- & \textbf{.695} & .733  & .011 &  .729 &  .821&  .024\\

SAM-DSA(ICCV'25)~\cite{SAMDSA} &\textbf{\MakeUppercase{\romannumeral 2}}
& .866 & - &.047 & .881 & -&.023 & .889  & -&.032 
&SAM-DSA(ICCV'25)~\cite{SAMDSA}& \textbf{\MakeUppercase{\romannumeral 2}}& .875  & - &.044 & .887 & -& .022 & .898 & -& \underline{.029} 
&EMIP(TIP'25)~\cite{emip}&-& .675 & .723 & .015& .719 & .819 & .028\\

VSCode-V2 (TPAMI'26)~\cite{vscode2} &\textbf{\MakeUppercase{\romannumeral 2}}
& .877 & \underline{.940} &\underline{.043} & .897 & \underline{.945}&\underline{.020} & .899  & \underline{.945}&.031 
&VSCode-V2 (TPAMI'26)~\cite{vscode2}& \textbf{\MakeUppercase{\romannumeral 2}}& \underline{.883} & \underline{.943} &\underline{.042} & \underline{.899} & \underline{.947}&\underline{.019} & .903  & \underline{.947}&\underline{.029}
&VSCode-V2 (TPAMI'26)~\cite{vscode2}&\textbf{\MakeUppercase{\romannumeral 2}}& .671 & \textbf{.791} & \textbf{.008}& \underline{.791} & \underline{.843} & \underline{.023}\\

\rowcolor[rgb]{0.918,0.965,0.973}
 Ours& 
 &\textbf{.897}  
 &\textbf{.945} 
 & \textbf{.039} 
 & \textbf{.902} 
 & \textbf{.958} 
 & \textbf{.017} 
 &\textbf{.912}
 &\textbf{.952} 
 &\textbf{.027}
& Ours & & \textbf{.899}  
& \textbf{.947} 
&\textbf{.038}
&\textbf{.902} 
&\textbf{.958} 
& \textbf{.017}
&\textbf{.911}
&\textbf{.952} 
&\textbf{.027}
 &Ours 
 & 
 &.646
 &\underline{.744}
 &  \underline{.010}
 & \textbf{.811} 
 &\textbf{.877} 
 &\textbf{.021}\\
\rowcolor[rgb]{0.918,0.965,0.973}
Average Improvement& &.037 &.026&.013&.022&.032&.007&.017&.031&.007
&Average Improvement&&.044&.103&.017&.036&.166&.010&.031&.118&.013&Average Improvement&&-.024&.031&.010&.107&.110&.010\\
\hline
\end{tabular}
\label{tab:I}
}
\vspace{0.8em} 

\resizebox{\linewidth}{!}{
    \begin{tabular}{r|c|cccc|cccc|r|c|cccc|r|c|cccc}
    \specialrule{1.5pt}{0pt}{0pt}
        \multicolumn{10}{c|}{\textbf{RGB MD}} & \multicolumn{6}{c|}{\textbf{RGB-D MD}} & \multicolumn{6}{c}{\textbf{VMD}}\\
        \midrule
        \multirow{2}{*}{Methods}
        &\multirow{2}{*}{Type}
        &\multicolumn{4}{c|}{MSD \cite{MSD}} 
        & \multicolumn{4}{c|}{PMD \cite{PMD}}
        & \multirow{2}{*}{Methods}
        &\multirow{2}{*}{Type}
        & \multicolumn{4}{c|}{RGBD-Mirror \cite{rgb-d-mirror}}
        & \multirow{2}{*}{Methods}
        &\multirow{2}{*}{Type}
        & \multicolumn{4}{c}{VMD-D \cite{VMD-D}} \\   &  
        & $S_m \uparrow$ 
        & $F_m \uparrow$ 
        & $E_m \uparrow$
        & $M \downarrow$
        & $S_m \uparrow$ 
        & $F_m \uparrow$ 
        & $E_m \uparrow$
        & $M \downarrow$
        & 
        &
        & $S_m \uparrow$ 
        & $F_m \uparrow$ 
        & $E_m \uparrow$
        & $M \downarrow$
        & 
        &
        & $S_m \uparrow$ 
        & $F_m \uparrow$ 
        & $E_m \uparrow$
        & $M \downarrow$
         \\
        \hline 
        UniMatch(CVPR'23) ~\cite{UniMatch}&\textbf{\MakeUppercase{\romannumeral 2}}& .921& .934& .951 & .031 & .868 & .835 & .916 & .028 
        & UniMatch(CVPR'23)~\cite{UniMatch} &\textbf{\MakeUppercase{\romannumeral 2}}& .868& .873& .912 & .041& VMD-Net(CVPR'23)~\cite{vmdnet} &\textbf{\MakeUppercase{\romannumeral 2}}&.789 
        & .691 & .794 & .105\\
        
        DepthAnything(CVPR'24) ~\cite{depthany}&\textbf{\MakeUppercase{\romannumeral 2}} & .931& .948& .964 & .025 & .873 & .841 & .917 & .024
        & DepthAnything(CVPR'24) ~\cite{depthany}&\textbf{\MakeUppercase{\romannumeral 2}} & .881 & .895 & .931 & .031 & DepthAnything(CVPR'24) ~\cite{depthany}&\textbf{\MakeUppercase{\romannumeral 2}}&.807 
        & .711 & .823 & .087 \\
        
        SAM2UNet(ICCV'25) ~\cite{sam2unet}&\textbf{\MakeUppercase{\romannumeral 1}}& .933& .947& .965 & .024 & .876 & .843 & .921 & .023
        & SAM2UNet(ICCV'25) ~\cite{sam2unet}&\textbf{\MakeUppercase{\romannumeral 1}}& .876&.882& .920 & .037  & SAM2UNet(ICCV'25) ~\cite{sam2unet}&\textbf{\MakeUppercase{\romannumeral 1}}& .805 & .713 & .827 & .086 \\
        
        DAM(CVPR'25)~\cite{DAM} &\textbf{\MakeUppercase{\romannumeral 2}}&  \underline{.941}&  \underline{.953}& \underline{.971} &  \underline{.019} &  \underline{.883} &  \underline{.846} &  \underline{.924} & \textbf{.020}
        
        & DAM(CVPR'25) ~\cite{DAM}&\textbf{\MakeUppercase{\romannumeral 2}}& \underline{.883}& \underline{.897}&  \underline{.933} &  \underline{.030}& DAM(CVPR'25) ~\cite{DAM} &\textbf{\MakeUppercase{\romannumeral 2}}&  \underline{.832} &  \underline{.732} & \textbf{.843} & \textbf{.075}
        \\
        
        \rowcolor[rgb]{0.918,0.965,0.973}
        Ours && \textbf{.946}& \textbf{.959}& \textbf{.975} & \textbf{.018} 
        & \textbf{.886} & \textbf{.849 }& \textbf{.925} & \textbf{.020}
        & Ours && \textbf{.886} &\textbf{.898}& \textbf{.936} & \textbf{.029}
        &Ours && \textbf{.835} & \textbf{.735} &  \underline{.841} &  \underline{.076}\\

        \rowcolor[rgb]{0.918,0.965,0.973}
Average Improvement& &.015 &.014&.013&.006&.011&.008&.006&.003&Average Improvement& &.009&.012&.012&.005&Average Improvement& &.027&.023&.020&.012\\
        \hline
    \end{tabular}
}
\vspace{0.8em} 

\resizebox{\linewidth}{!}{
    \begin{tabular}{r|c|cccc|cccc|r|c|cccc|r|c|cc|r|c|cc}
    \specialrule{1.5pt}{0pt}{0pt}
        \multicolumn{10}{c|}{\textbf{MAS}} & \multicolumn{6}{c|}{\textbf{SD}} 
        & \multicolumn{4}{c|}{\textbf{TOS}}
        & \multicolumn{4}{c}{\textbf{DBD}}\\
        \midrule
        \multirow{2}{*}{Methods}
        &\multirow{2}{*}{Type}
        &\multicolumn{4}{c|}{MAS3K \cite{MAS3K}} 
        & \multicolumn{4}{c|}{RMAS \cite{RMAS}}
        & \multirow{2}{*}{Methods}
        &\multirow{2}{*}{Type}
        & \multicolumn{4}{c|}{SBU \cite{SBU}}
        & \multirow{2}{*}{Methods}
        &\multirow{2}{*}{Type}
        & \multicolumn{2}{c|}{Trans 10K \cite{Transparent10k}}
        &\multirow{2}{*}{Methods}
        &\multirow{2}{*}{Type}
        & \multicolumn{2}{c}{CHUK \cite{CUHK}} \\ 
        &  
        & $S_m \uparrow$ 
        & $F_m \uparrow$ 
        & $E_m \uparrow$
        & $M \downarrow$
        & $S_m \uparrow$ 
        & $F_m \uparrow$ 
        & $E_m \uparrow$
        & $M \downarrow$
        & 
        &
        & $Iou \uparrow$ 
        & $Dice \uparrow$ 
        & $F_m^w \uparrow$
        & $M \downarrow$
        & 
        &
        & $BER \downarrow$
        & $M \downarrow$
        & 
        &
        & $F_{\beta} \uparrow$
        & $M \downarrow$
         \\
        \hline 
        EVP(CVPR'23)~\cite{EVP}& \textbf{\MakeUppercase{\romannumeral 1}}& .881 & .870 & .928 & .024 & .858 & .849& .939 & .024 
        & EVP(CVPR'23)~\cite{EVP} &\textbf{\MakeUppercase{\romannumeral 1}}& .815& .853& .809 & .038
        & RFENet(IJCAI’23)~\cite{rfenet} & -& .074
        & .115 
        &DD(ECCV’20)~\cite{DD}&-& .928 & .042\\
        
        Dual-SAM(CVPR'24)~\cite{Dual-sam} &-& .884 & .873 & .933 & .023 & .860 & .852& .944 & .022
        & FSEL(ECCV'24) ~\cite{FSEL}&\textbf{\MakeUppercase{\romannumeral 1}} & .835& .893& .868 & .027 
        
        & EVP(CVPR'23) ~\cite{EVP}&\textbf{\MakeUppercase{\romannumeral 1}}& .059 & .069 
        & EVP(CVPR'23) ~\cite{EVP}&\textbf{\MakeUppercase{\romannumeral 1}} &.928 & .045 \\ 
        SAM2UNet(ICCVW'25)~\cite{sam2unet}&\textbf{\MakeUppercase{\romannumeral 1}}& .903& .891 & .943 &  \underline{.021} & .874 & .866 & .944 & .022
        & SAM2UNet(ICCV'25)~\cite{sam2unet}& \textbf{\MakeUppercase{\romannumeral 1}}&.851& .903& .883 & .025  
        & Spider(ICML'24) ~\cite{spider}&\textbf{\MakeUppercase{\romannumeral 1}}&  \underline{.050} &  \underline{.053} & Spider(ICML'24) \cite{spider} & \textbf{\MakeUppercase{\romannumeral 1}}& .931 & .038 \\

        FOCUS(AAAI'25) ~\cite{focus}&\textbf{\MakeUppercase{\romannumeral 1}}& .906 & .895 & .950 & .024& .879 & .871& .951 & .023
        & Spider(ICML'24) ~\cite{spider}&\textbf{\MakeUppercase{\romannumeral 1}}& .823&.893& .868 & .027
        & FOCUS(AAAI'25) ~\cite{focus}&\textbf{\MakeUppercase{\romannumeral 1}}& .054 & .057 & FOCUS(AAAI'25) \cite{focus} & \textbf{\MakeUppercase{\romannumeral 1}}&  \underline{.934} &  \underline{.036}\\
        
        SAM2-BGNet(SJ'25)\cite{SAM2BGNET} &-&  \underline{.907} &  \underline{.897} &  \underline{.952} & .022&  \underline{.880} &  \underline{.872} & \underline{.952} &  \underline{.022}
        & LPMoE(ICCV'25) ~\cite{DAM}&\textbf{\MakeUppercase{\romannumeral 1}}&  \underline{.861}& \underline{.917}&  \underline{.912} &  \underline{.022}
        & SAM2-UNet(ICCVW'25)~\cite{sam2unet} &\textbf{\MakeUppercase{\romannumeral 1}}& .051 & .055& SAM2-UNet(ICCVW'25) \cite{sam2unet} &\textbf{\MakeUppercase{\romannumeral 1}}& .928 & .044
        \\ 
        \rowcolor[rgb]{0.918,0.965,0.973}
        Ours &
        & \textbf{.910}
        & \textbf{.900}
        & \textbf{.956} 
        & \textbf{.020} 
        & \textbf{.884} 
        & \textbf{.877} 
        & \textbf{.957} 
        & \textbf{.020}
        
        & Ours && \textbf{.874}& \textbf{.929}&\textbf{.926} & \textbf{.019}&
        Ours && \textbf{046} &\textbf{.049} &Ours&& \textbf{.936} & \textbf{.035}\\

        \rowcolor[rgb]{0.918,0.965,0.973}
Average Improvement& &.014 &.015&.015&.002&.014&.015&.011&.002&Average Improvement& &.037&.038&.058&.008&Average Improvement& &.011&.020&Average Improvement& &.007&.006\\
        \hline
    \end{tabular}
}
\label{tab:1}
\end{table*}
\begin{table}[t]
    \centering 
    \caption{Comparison of DifferSeg with recent state-of-the-art methods in medical image segment. The best scores are \textbf{bold}.}
    \setlength{\tabcolsep}{0.4mm} 
\resizebox{\linewidth}{!}{
\begin{tabular}{r|c|cc|cc|cc|cc}
\toprule 
\multirow{2}{*}{Methods}
&\multirow{2}{*}{Type}

& \multicolumn{2}{c|}{\textbf{CPS}} 
& \multicolumn{2}{c|}{\textbf{CLI}}
& \multicolumn{2}{c|}{\textbf{BLS}} 
& \multicolumn{2}{c}{\textbf{SLS}} \\

&
&mDice$\uparrow$  & mIoU$\uparrow$
&mDice$\uparrow$  & mIoU$\uparrow$
&mDice$\uparrow$  & mIoU$\uparrow$
&mDice$\uparrow$  & mIoU$\uparrow$
\\ \hline
CTNet(M)(CVPR'25)~\cite{CTNet-M}&- & .851 &.781 & - &- & - & - & - & - \\
DECOR-Net(ISBI'23)~\cite{DECOR-Net}&-& - &-& .639 & .498  & -&- &- & -  \\
DBUNet(BSPC'25)~\cite{DBUNet} &-& - &- & -&-& .797 & .725 &- & -  \\

FlowSDF(IJCV'25)~\cite{flowsdf}&- & - &-  & - &- & -& -&.894 & .824 \\
Spider(ICML'24)~\cite{spider}&\textbf{\MakeUppercase{\romannumeral 1}}& .827 &.742 & .648 &.579& .798 & .727 &.867 & .817  \\
SAM2UNet(ICCVW'25)~\cite{sam2unet}&\textbf{\MakeUppercase{\romannumeral 1}}& .847 &.786& .645 &.583& .831 & .748 &.891 & .821  \\
\rowcolor[rgb]{0.918,0.965,0.973}
 Ours &-& \textbf{.854} & \textbf{.795} & \textbf{.666} &\textbf{.595}& \textbf{.837} &\textbf{.759} & \textbf{.899} & \textbf{.830}\\

 \rowcolor[rgb]{0.918,0.965,0.973}
Average Improvement& &.013 &.026&.022&.042&.029&.026&.015&.010\\
\hline
\end{tabular}
\label{tab:2}
}
\end{table}
\textbf{Datasets.}
For the SOD tasks, we selected three widely used benchmark datasets for evaluation in each of the four modalities: \textbf{RGB SOD}: DUTS \cite{DUTS}, DUT-O \cite{DUT-O}, and HKU-IS \cite{HKU-IS}; \textbf{RGB-D SOD}: STERE \cite{STERE}, NLPR \cite{NLPR}, and DUTLF \cite{DUT-Depth}; \textbf{RGB-T SOD}: VT821 \cite{VT821}, VT1000 \cite{VT1000}, and VT5000 \cite{VT5000}; \textbf{VSOD}: FBMS\cite{FBMS}, SegV2 \cite{sam-cod}, and DAVSOD-easy \cite{DAVSOD}. For the \textbf{COD} tasks, we evaluated DifferSeg on three large-scale benchmark datasets under the RGB and RGB-D modalities, namely CAMO \cite{CAMO}, COD10K \cite{COD10K}, and NC4K \cite{NC4K}.
For the \textbf{VCOD} task, we employed two widely recognized benchmark datasets: CAD \cite{CAD} and MoCA-Mask \cite{MoCA-Mask}. In the \textbf{MD} tasks, we also selected representative datasets for the three modality-specific sub-tasks, including MSD \cite{MSD}, PMD \cite{PMD}, RGBD-Mirror \cite{rgb-d-mirror}, and VMD-D \cite{VMD-D}. For the \textbf{MAS} task, two large-scale benchmark datasets, MAS3K \cite{MAS3K} and RMAS \cite{RMAS}, were used for evaluation. In the \textbf{DBD}, \textbf{SD}, and \textbf{TOS} tasks, we used CUHK \cite{CUHK}, SBU \cite{SBU}, and Transparent10K\cite{Transparent10k} for evaluation, respectively. Finally, for the four medical image segmentation tasks, we followed previous works~\cite{CTNet-M,spider} by combining five datasets \cite{colonDB,cvc300,ClinicDB,kvasir,etis} for comprehensive evaluation in the \textbf{CPS} task, while the remaining three medical tasks were validated on the COVID-19\cite{COVID}, BUSI\cite{BUSI}, and ISIC2018\cite{ISIC} datasets. 

\textbf{Evaluation metrics.}
During evaluation, we followed the standard practices established in each research domain and adopted widely used metrics to ensure fair and comprehensive comparison with prior works. For the main tasks of salient object detection (SOD) and camouflaged object detection (COD), which are the primary focus of many related studies, we report four commonly used metrics: structure-measure ($S_m$)~\cite{SA}, weighted F-measure ($F_m$)~\cite{Fmeasure}, max enhanced alignment measure ($E_m$)~\cite{EMAX}, and mean absolute error ($M$)~\cite{Mae}. These metrics collectively evaluate structural similarity, region-level accuracy, global alignment, and pixel-wise error.

Given the substantial differences in task characteristics and evaluation conventions across domains, enforcing a single cross-task unified evaluation protocol is not appropriate. Therefore, we adopt domain-specific metrics for each sub-task, strictly following the established benchmarks and evaluation practices in the corresponding literature.

For shadow detection (SD), we report Intersection over Union (IoU), Dice coefficient, weighted F-measure ($F_m^w$), and mean absolute error ($M$).
For transparent object segmentation (TOS), we use balance error rate (BER)~\cite{ber} and mean absolute error ($M$).
For defocus blur detection (DBD), we report weighted F-measure ($F_{\beta}$) and mean absolute error ($M$).
For medical image segmentation, performance is evaluated using mean Dice coefficient (mDice) and mean IoU (mIoU).

Higher values indicate better performance for all metrics except $M$ and BER, where lower values are preferred. This domain-specific evaluation strategy ensures that our results are directly comparable to state-of-the-art methods within each respective field.

\textbf{Implementation details.}
Our DifferSeg model is implemented based on PyTorch \cite{pytorch} and trained on an NVIDIA RTX A800 GPU with 80GB of memory. For optimization, we use the Adam \cite{adam} optimizer with an initial learning rate of 5e-3. The model is trained for 30 epochs on natural image tasks and 20 epochs on medical image tasks until convergence. For backbone initialization, we use the Hiera-L version of SAM2 \cite{SAM2} pretrained weights to accelerate training and improve performance. 

We follow previous works~\cite{spider,DAM,vscode,sam2unet,focus} and use the same training datasets for each corresponding task. Specifically, the DUTS~\cite{DUTS} training set is used for \textbf{RGB SOD}. The NJUD~\cite{NJU2K}, NLPR~\cite{NLPR}, and DUTLF-Depth~\cite{DUT-Depth} training sets are used for \textbf{RGB-D SOD}. The VT5000~\cite{VT5000} training set is used for \textbf{RGB-T SOD}. The DAVIS~\cite{DAVIS}, FBMS~\cite{FBMS}, and DAVSOD~\cite{DAVSOD} training sets are used for \textbf{VSOD}. The COD10K~\cite{COD10K} and CAMO~\cite{CAMO} training sets are used for both \textbf{RGB COD} and \textbf{RGB-D COD}, and the MoCA-Mask~\cite{MoCA-Mask} training set is used for \textbf{VCOD}.
For \textbf{MAS}, we use the RMAS~\cite{RMAS} and MAS3K~\cite{MAS3K} training sets, while the MSD~\cite{MSD} and PMD~\cite{PMD} training sets are used for \textbf{MD}. The RGB-D Mirror~\cite{rgb-d-mirror} training set is used for \textbf{RGB-D MD}, and the VMD-D~\cite{VMD-D} training set is used for \textbf{VMD}. The ISTD~\cite{ISTD} training set is used for \textbf{SD}. The CHUK~\cite{CUHK} and DUT~\cite{DUT} training sets are used for \textbf{DBD}. The Transparent10K~\cite{Transparent10k} training set is used for \textbf{TOS}.
Five polyp segmentation datasets~\cite{ClinicDB,colonDB,etis,kvasir,cvc300} are used for \textbf{CPS}. The COVID-19~\cite{COVID} dataset is used for \textbf{CLI}. The BUSI~\cite{BUSI} dataset is used for \textbf{BLS}, and the ISIC2018~\cite{ISIC} dataset is used for \textbf{SLS}. For the VSOD, VMD, and VCOD tasks, we follow common practice and use FlowNet2.0 \cite{flownet} as the optical flow extractor due to its strong performance. All tasks are evaluated using a unified PyTorch-based toolbox.

\begin{figure*}[!t]
    \centering
    \includegraphics[width=0.95\linewidth]{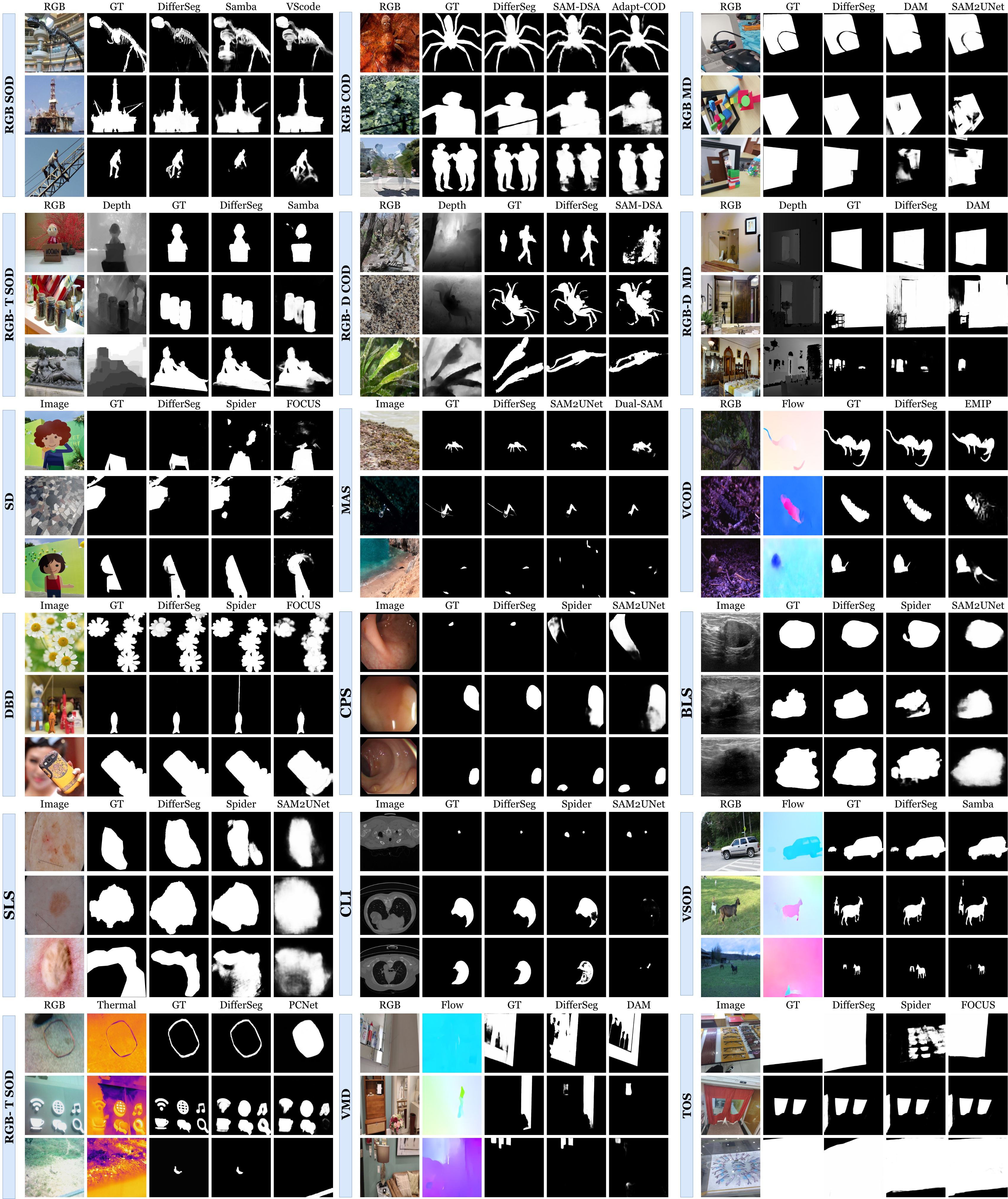}
    \caption{Qualitative comparison of DifferSeg with state-of-the-art methods across multiple binary segmentation tasks. (GT: ground truth.)}
    \label{fig:image6}
\end{figure*}
\subsection{Comparison with State-of-the-Art Methods}
To comprehensively evaluate the generality and effectiveness of the proposed method, DifferSeg is systematically compared with the latest \textit{General \MakeUppercase{\romannumeral 1}} models, \textit{General \MakeUppercase{\romannumeral 2}} models, and multiple task-specific approaches in the field of binary segmentation. The evaluation covers 18 downstream binary segmentation tasks, spanning a wide range of representative applications in both natural and medical scenarios. 

\textbf{Quantitative Comparison.}
In Tab.~\hyperref[tab:1]{1}, we conduct a large-scale evaluation of DifferSeg against recent state-of-the-art specialized and general models on fourteen natural image segmentation tasks. On camouflaged object detection benchmarks, DifferSeg demonstrates clear and consistent advantages over existing methods, achieving improvements of over +1\% in both $S_m$ and $E_m$ on challenging datasets such as CAMO, COD10K, and NC4K. These results highlight the superior capability of DifferSeg in segmenting low-contrast and visually confusing targets. For other natural image segmentation tasks, including RGB/RGB-D/RGB-T salient object detection, mirror detection, marine animal segmentation, shadow detection, transparent object segmentation, and defocus blur detection, DifferSeg consistently ranks among the top-performing methods across all evaluation metrics.

In Tab.~\hyperref[tab:2]{2}, we further compare DifferSeg on four medical image segmentation tasks, including CPS, CLI, BLS, and SLS. DifferSeg achieves the best performance across all benchmarks. Notably, on the challenging CLI dataset, DifferSeg outperforms the strongest competing general model by +2.1\% in mDice and +1.2\% in mIoU, indicating a clear advantage in handling ambiguous boundaries and complex anatomical structures. Consistent improvements are also observed on the remaining medical datasets, demonstrating the robustness of DifferSeg across diverse medical imaging modalities.

It is worth noting that the performance margin on CPS is relatively modest compared with the strongest existing method. Specifically, DifferSeg improves the mDice score from 0.851 to 0.854 compared with CTNet. This suggests that when existing methods already achieve strong performance on certain datasets, further improvement over the current SOTA can be limited. Nevertheless, DifferSeg still achieves the best CPS results and shows consistent overall performance across the four medical segmentation tasks.
We also acknowledge that DifferSeg introduces additional computational cost due to the use of a relatively large backbone. Therefore, the advantage of DifferSeg is more evident when overall performance across multiple datasets and robustness across different medical segmentation tasks are prioritized. In contrast, for resource-constrained deployment scenarios, the accuracy-efficiency trade-off should be carefully considered, especially when the gain over the strongest existing method on a specific dataset is relatively small. Nevertheless, DifferSeg still achieves the best CPS results and shows consistent overall performance across the four medical segmentation tasks.

Overall, the consistent performance gains observed on challenging benchmarks, together with strong stability across 18 downstream tasks, demonstrate that DifferSeg effectively surpasses existing specialized methods as well as General I and General II models, validating its effectiveness as a general binary segmentation framework.

\textbf{Qualitative Comparison.}
We present visual comparisons of DifferSeg against other state-of-the-art methods across 18 binary segmentation tasks. Fig.~\hyperref[fig:image6]{6} presents qualitative results on several representative tasks, including four salient object detection (SOD) tasks, three camouflaged object detection (COD) tasks, three mirror detection (MD) tasks, as well as marine animal segmentation (MAS), transparent object segmentation (TOS), shadow detection (SD), defocus blur detection (DBD), and four medical image segmentation tasks.

From the visual results, it can be observed that DifferSeg achieves stable and strong segmentation performance across a variety of challenging scenarios. The proposed method effectively handles objects of extremely small or large scales, manages multiple or occluded targets, and maintains accurate segmentation when object boundaries are ambiguous or blurry. In contrast, existing methods often suffer from missed detections or inaccurate boundaries in these scenarios, while DifferSeg demonstrates stronger robustness.

\subsection{Parametric Analysis}
\begin{table}[t]
    \centering
    \caption{Comprehensive cost analysis of the proposed general framework.}
    \setlength{\tabcolsep}{0.5mm} 
    \resizebox{\linewidth}{!}{%
    \begin{tabular}{c|cccccccc}
    \specialrule{1.5pt}{0pt}{0pt}
    
    \toprule
    Methods& GateNet& VSCode & Spider &VSCode-V2& SAM2UNet &DifferSeg & DPF &FGD \\
    \hline
Learnable Params &92.8 M& 74.7 M & 89.2 M &90.3 M& 53.9 M& 85.0 M& 26.2 M&27.4 M\\
FLOPs &94.1 G& 89.4 G& -&96.4 G & 292.4 G& 322.2 G& 8.1 G& 21.4 G \\
FPS &29.5& 20.4 & - &19.1 & 10.3 & 10.0 & -& - \\
    \specialrule{1.5pt}{0pt}{0pt}
    \end{tabular}%
    }
    \label{tab:3}
\end{table}
Tab.~\hyperref[tab:3]{3} provides a systematic comparison of several representative methods in terms of learnable parameter scale and computational complexity. As shown, although DifferSeg adopts a multi-module design, its overall number of learnable parameters is 85.0M, which is still significantly lower than that of the \textit{General \MakeUppercase{\romannumeral 1}} methods GateNet (92.8M) and Spider (89.2M), and remains at a comparable level to the \textit{General \MakeUppercase{\romannumeral 2}} method VSCode (74.7M), demonstrating good parameter efficiency. In terms of computational cost, DifferSeg has a total complexity of 322.2G FLOPs, which mainly originates from the inherent computation of the backbone. It is worth noting that the two core modules introduced in DifferSeg, namely DPF and FGD, remain highly lightweight in both parameter scale and computational cost. Specifically, DPF introduces only 26.2M parameters and 8.1G FLOPs, while FGD requires merely 27.4M parameters and 21.4G FLOPs. While significantly enhancing feature representation and task decoupling capability, these modules impose a limited impact on the overall computational cost.

Overall, DifferSeg is designed primarily as a general multimodal binary segmentation framework rather than an edge-optimized model. Although its measured FPS is comparable to SAM2UNet on
high-end GPU hardware, its computational cost remains relatively high for latency-sensitive
edge deployment. Developing lightweight variants with smaller SAM2 backbones, pruning,
quantization, or distillation is an important future direction.
\subsection{Ablation studies}
\begin{table*}[t]
    \centering 
    \caption{Component-wise ablation study of DifferSeg across different task datasets. The best scores are highlighted in \textbf{Bold}.}
\resizebox{1\linewidth}{!}{
\begin{tabular}{c|ccc|ccc|ccc|ccc|ccc|ccc|ccc|ccc}
\toprule
\multirow{3}{*}{Settings}
& \multicolumn{3}{c|}{\textbf{RGB SOD}} 
& \multicolumn{3}{c|}{\textbf{RGB COD}} 
& \multicolumn{3}{c|}{\textbf{RGB MD}} 
& \multicolumn{3}{c|}{\textbf{RGB-D SOD}}
& \multicolumn{3}{c|}{\textbf{RGB-D COD}}
& \multicolumn{3}{c|}{\textbf{VSOD}} 
& \multicolumn{3}{c|}{\textbf{VCOD}} 
& \multicolumn{3}{c}{\textbf{RGB-D MD}}  \\

& \multicolumn{3}{c|}{DUT-O} 
& \multicolumn{3}{c|}{CAMO}
& \multicolumn{3}{c|}{MSD} 
& \multicolumn{3}{c|}{STERE} 
& \multicolumn{3}{c|}{CAMO}
& \multicolumn{3}{c|}{FBMS}
& \multicolumn{3}{c|}{CAD} 
& \multicolumn{3}{c}{RGB-D Mirror} \\

& $S_m \uparrow$  & $E_m \uparrow$ & $M \downarrow$
& $S_m \uparrow$  & $E_m \uparrow$ & $M \downarrow$
& $S_m \uparrow$  & $E_m \uparrow$ & $M \downarrow$
& $S_m \uparrow$  & $E_m \uparrow$ & $M \downarrow$
& $S_m \uparrow$  & $E_m \uparrow$ & $M \downarrow$
& $S_m \uparrow$  & $E_m \uparrow$ & $M \downarrow$
& $S_m \uparrow$  & $E_m \uparrow$ & $M \downarrow$
& $S_m \uparrow$  & $E_m \uparrow$ & $M \downarrow$
\\
\hline 
Baseline& .884&.881&.038&.885&.931 &.042 & .931 & .963 & .025 &.926&.944 &.030 &.883&.930&.043&.911&.930&.028& .746&.783 & .030 & .877&.921& .035  \\
w/o DPF& .895 &.926& .033& .897 &.945 & .039 & .946 & .975&.018&.932&.949&.023 &.891&.942&.040&.918&.940&.024&.771&.826&.025&.882&.928&.032 \\
w/o FGD&.884 &.881& .038& .885 &.931& .042 & .931 & .963 &.025 & .930&.947 & .027&.893&.940&.041&.922&.943&.023& .766&.808 &.026 &.881&.927& .033 \\
w/o Adapter& .812& .804 & .072 & .802 &.856 & .061 &.863& .872&.049 & .843&.851 & .053&.808&.851&.064&.783&.800&.500& .657&.703& .047 &.748&.807& .057 \\
\rowcolor[rgb]{0.918,0.965,0.973}Full (DifferSeg)&\textbf{.895}&\textbf{.926} & \textbf{.033} &\textbf{.897}&\textbf{.945} & \textbf{.039} &\textbf{.946} & \textbf{.975} &\textbf{.018} & \textbf{.939}&\textbf{.966} & \textbf{.023}&\textbf{.899}&\textbf{.947} &\textbf{.038} &\textbf{.930}&\textbf{.957}&\textbf{.021}& \textbf{.811}&\textbf{.877}& \textbf{.021}&   \textbf{.886}&\textbf{.936} & \textbf{.029}\\ 
\hline

\end{tabular}
\label{tab:4}
}
\end{table*}
\begin{table}[t]
    \centering 
    \caption{Ablation study of the DPF module across different task datasets. The best scores are highlighted in \textbf{Bold}.}
\resizebox{\linewidth}{!}{
\begin{tabular}{c|ccc|ccc|ccc|ccc}
\toprule
\multirow{3}{*}{Settings}
& \multicolumn{3}{c|}{\textbf{RGB-D SOD}} 
& \multicolumn{3}{c|}{\textbf{RGB-D COD}} 
& \multicolumn{3}{c|}{\textbf{RGB-D MD}} 
& \multicolumn{3}{c}{\textbf{VCOD}}
\\

& \multicolumn{3}{c|}{STERE} 
& \multicolumn{3}{c|}{CAMO}
& \multicolumn{3}{c|}{RGB-D Mirror} 
& \multicolumn{3}{c}{CAD} 
\\
& $S_m \uparrow$  & $E_m \uparrow$ & $M \downarrow$
& $S_m \uparrow$  & $E_m \uparrow$ & $M \downarrow$
& $S_m \uparrow$  & $E_m \uparrow$ & $M \downarrow$
& $S_m \uparrow$  & $E_m \uparrow$ & $M \downarrow$
\\
\hline 
MMFM&.935 & .959 & .025 & .892 & .942& .041& .879 & .932 &0.33&.803&.851&.022\\
3x3 Learn Conv& .938&.962 & .024&.894&.944&.039 &.882 &.934&. 030& .787 &.849& .025 \\
w/o Low-frequency& .935&.959 & .025&.893&.942&.041 &.880 &.931&. 031& .787 &.849& .025  \\
Fixed Differential Operators& .937&.962 & .024&.895&.943&.040 &.882 &.932&. 030& .796 &.854& .024  \\
\rowcolor[rgb]{0.918,0.965,0.973}DPF&\textbf{.939} & \textbf{.966} & \textbf{.023}&\textbf{.899}&\textbf{.947}&\textbf{.038} & \textbf{.886}& \textbf{.936} & \textbf{.029} & \textbf{.811}&\textbf{.877}&\textbf{.021}\\ 
\hline 
\end{tabular}
\label{tab:5}
}
\end{table}
\begin{table}[t]
    \centering 
    \caption{Ablation study of the FGD module across different task datasets. The best scores are highlighted in \textbf{Bold}.}
\resizebox{\linewidth}{!}{
\begin{tabular}{c|ccc|ccc|ccc|ccc}
\toprule
\multirow{3}{*}{Settings}
& \multicolumn{3}{c|}{\textbf{RGB SOD}} 
& \multicolumn{3}{c|}{\textbf{RGB COD}} 
& \multicolumn{3}{c|}{\textbf{RGB-D MD}} 
& \multicolumn{3}{c}{\textbf{VCOD}}
\\

& \multicolumn{3}{c|}{DUT-O} 
& \multicolumn{3}{c|}{CAMO}
& \multicolumn{3}{c|}{RGB-D Mirror} 
& \multicolumn{3}{c}{CAD} 
\\

& $S_m \uparrow$  & $E_m \uparrow$ & $M \downarrow$
& $S_m \uparrow$  & $E_m \uparrow$ & $M \downarrow$
& $S_m \uparrow$  & $E_m \uparrow$ & $M \downarrow$
& $S_m \uparrow$  & $E_m \uparrow$ & $M \downarrow$
\\
\hline 
 w/o SA& .892 &.923& .034& .890 &.934 & .041 &.883& .932&.030 & .795 & .852& .024\\
w/o FB& .892 &.922& .034& .894&.939 & .040 & .878 &.930& .031 & .794&.866& .023 \\
Single-Path&.889 &.229 &.036 & .894 &.941 & .040 &.882 &.931 & .031 &.794&.855&.023 \\
Transformer-Based& .889&.920 & .035& .892&.936& .041 & .879 &.931& .032 & .798&.857 & .024  \\
\rowcolor[rgb]{0.918,0.965,0.973}FGD&\textbf{.895} & \textbf{.926} & \textbf{.033} & \textbf{.899} & \textbf{.947}& \textbf{.038} & \textbf{.886} & \textbf{.936}& \textbf{.029}& \textbf{.811}& \textbf{.877}& \textbf{.021}\\ 
\hline 
\end{tabular}
\label{tab:6}
}
\end{table}
\begin{table}[h]
    \centering 
    \caption{Ablation study of DifferSeg in medical image segmentation tasks. The best scores are highlighted in \textbf{Bold}.}
\resizebox{\linewidth}{!}{
\begin{tabular}{c|cc|cc|cc|cc}
\toprule
\multirow{2}{*}{Settings}
& \multicolumn{2}{c|}{\textbf{CPS}} 
& \multicolumn{2}{c|}{\textbf{CLI}} 
& \multicolumn{2}{c|}{\textbf{BLS}}
& \multicolumn{2}{c}{\textbf{SLS}}
\\
&mDice$\uparrow$  & mIoU$\uparrow$
&mDice$\uparrow$  & mIoU$\uparrow$
&mDice$\uparrow$  & mIoU$\uparrow$
&mDice$\uparrow$  & mIoU$\uparrow$
\\
\hline 
Baseline& .847 & .785& .643  & .581 & .830&.749&.890&.821   \\
w/o SA& .853 & .794& .663  & .593 & .835&.756&.896&.826  \\
w/o FB& .852 & .793& .663  & .594 & .836&.755&.896&.827  \\
Single-Path& .851 & .792& .662  & .592 & .833&.752&.895&.826  \\
w/o Adapter& .731 & .658& .427  & .399 & .684&.537& .784&.709\\
\rowcolor[rgb]{0.918,0.965,0.973}DifferSeg&\textbf{.854} & \textbf{.795} & \textbf{.666} & \textbf{.595} & \textbf{.837}& \textbf{.759} & \textbf{.899} & \textbf{.830}\\ 
\hline 
\end{tabular}
\label{tab:7}
}
\end{table}
\begin{figure}[t]
    \centering
    \includegraphics[width=\columnwidth]{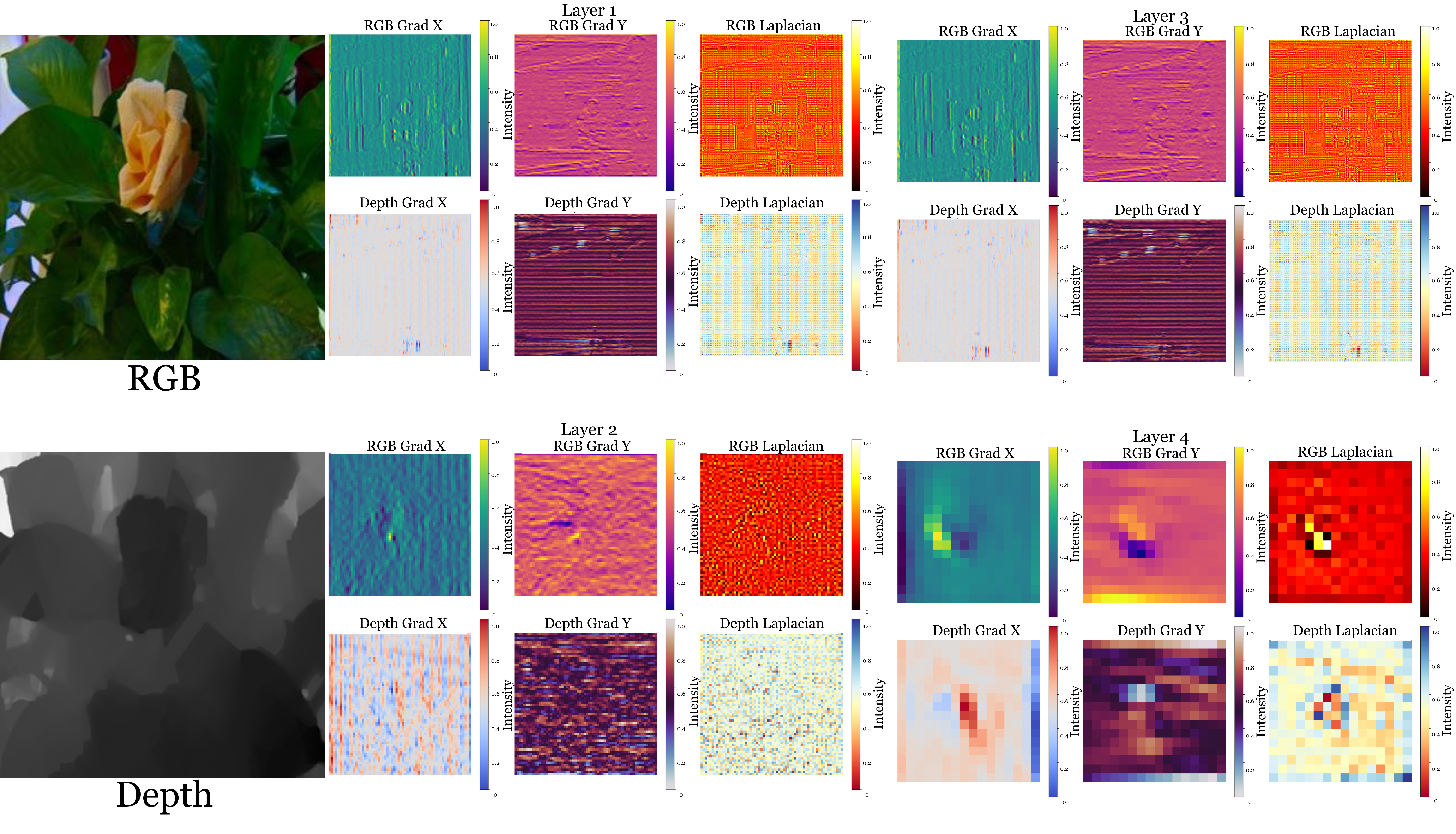}
    \caption{Visualization of different operators across different modalities and hierarchical levels. Within each set of images, each modality employs three learnable operators at four hierarchical levels to extract the corresponding features.}
    \label{fig:image7}
\end{figure}
\begin{figure}[t]
    \centering
    \includegraphics[width=\columnwidth]{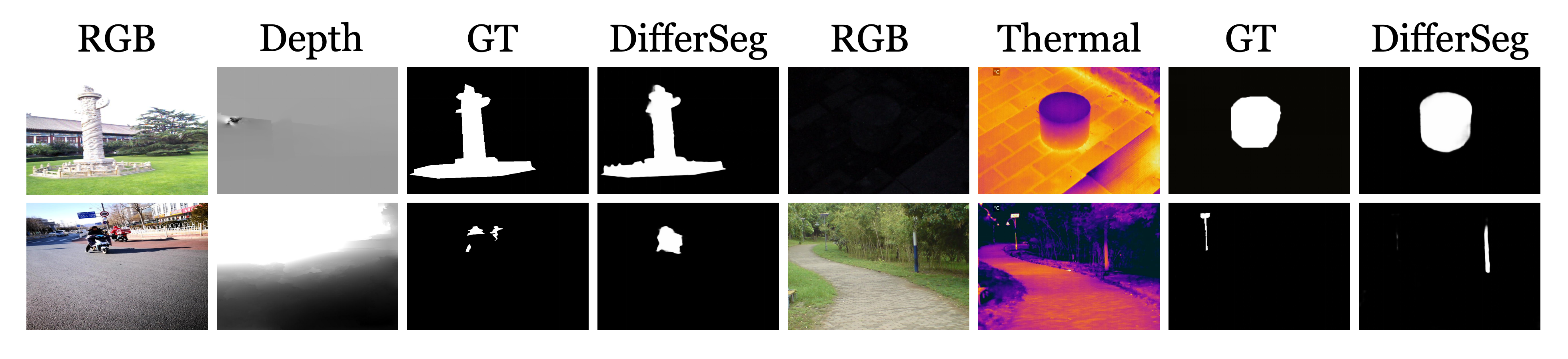}
\caption{Qualitative examples under modality failure cases.}
\label{fig:image8}
\end{figure}
\textbf{Component-wise Ablation.}
Tab.~\hyperref[tab:4]{4} reports the ablation results of the proposed components. We first constructed a baseline model that includes the SAM2 encoder, element-wise addition for fusion, and a U-shaped skip connection decoder. Based on this baseline, we progressively removed DPF, FGD, and the Adapter from DifferSeg for validation. The results compared to Full (DifferSeg) indicate that both DPF and FGD provide substantial performance gains over simple fusion and the traditional decoder used in the baseline. DPF effectively addresses modality mismatch and fusion difficulties, while FGD mitigates the loss of high-frequency details and the influence of low-frequency noise commonly observed in conventional decoding. We also evaluated the role of the Adapter in downstream tasks, and the results show that its inclusion enables the model to adapt effectively across diverse tasks.

\begin{figure}[t]
    \centering
    \includegraphics[width=\columnwidth]{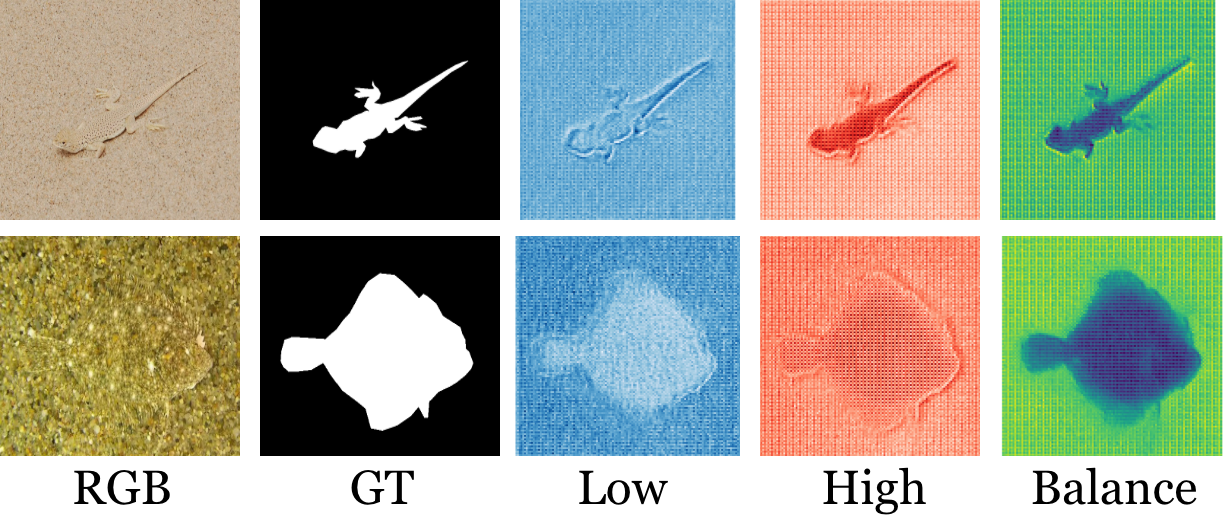}
\caption{Visualization of frequency feature balance. Low denotes low-frequency features, High denotes high-frequency features, and Balance denotes the fused balanced features.}

    \label{fig:image9}
\end{figure}

\textbf{Effectiveness of DPF Module.}
To evaluate the effectiveness of the differential perception fusion (DPF) module, we conducted four comparative experiments. First, we replaced the learnable differential operators with fixed differential operators. Second, we substituted all differential operators with 3×3 learnable convolutions. Third, we removed the low-frequency supplementation module. The comparison results indicate that the learnable differential operators can adaptively align and enhance cross-modality complementarity, while the low-frequency supplementation preserves global semantic consistency.
Beyond these internal comparisons, we also introduced the latest cross-modal fusion approach (MMFM) \cite{samba} for evaluation. As reported in Tab.~\hyperref[tab:5]{5}, our method achieves consistently superior performance across all downstream tasks compared with MMFM. This advantage primarily comes from the learnable differential operators, which enhance cross-modality complementarity through pixel-level complementary modeling and residual fusion, effectively mitigating modality mismatch and fusion challenges. 

In addition, Fig.~\hyperref[fig:image7]{7} and Fig.~\hyperref[fig:image8]{8} further supplement the evidence for the effectiveness of DPF.

Specifically, Fig.~\hyperref[fig:image7]{7} visualizes different operators across different modalities and hierarchical levels. Within each group of images, each modality employs three learnable operators at four hierarchical levels to extract the corresponding structured features. These visualizations show that the learned operators can capture modality-specific gradient, edge, and structural responses, supporting the effectiveness of the differential-inspired design in DPF. Fig.~\hyperref[fig:image8]{8} presents qualitative examples under modality failure cases. In the first row, the depth map or RGB image becomes unreliable. In these scenarios, DPF can suppress the ineffective modality and still produce reasonable predictions. However, as shown in the second row, when the thermal image or depth map fails under more challenging conditions, fusion redundancy may still exist due to extreme modality gaps. These examples demonstrate that DPF can effectively alleviate the negative impact of modality failure in some cases, while also revealing its limitation under severe modality degradation or weak cross-modal alignment.

\textbf{Effectiveness of FGD Module.}
To show the effectiveness of the FGD module, we design four comparative experiments. First, we remove the scale attention (SA) module from the original framework. Second, we remove the Frequency Balance (FB) module. Third, we replace the multi-path decoding strategy with a single-path strategy. Finally, we compare our decoder with a Transformer-based decoding approach~\cite{vst}. As shown in Tab.~\hyperref[tab:6]{6}, the SA module plays a key role in promoting interactions between low-frequency and high-frequency features and in enhancing overall feature representation, while the FB module enables more effective fusion by preserving global structures and fine-grained details. Furthermore, Fig.~\hyperref[fig:image9]{9} provides visualizations that demonstrate the balancing behavior within the FB module.
Compared with Transformer-based decoders and U-shaped decoders, FGD exhibits stronger adaptability across diverse downstream tasks and achieves consistently superior performance. This improvement stems from its ability to prevent the suppression of high-frequency details and the corruption of low-frequency components that often occur in traditional decoders, thereby establishing a more general and effective decoding paradigm for binary segmentation tasks.

\textbf{Effectiveness on Medical Image Segmentation.}
Tab.~\hyperref[tab:7]{7} presents the ablation study of DifferSeg on medical image segmentation tasks. First, we remove the Scale Attention (SA) module from the original framework. Second, we remove the Frequency Balance (FB) module. Third, we replace the multi-path decoding strategy with a single-path version. Finally, we remove the Adapter used for downstream task adaptation.

The experimental results clearly indicate that the SA module is crucial for enabling effective interactions between low-frequency and high-frequency features, thereby enhancing the overall feature representation. The FB module further improves feature fusion by jointly preserving global structures and fine-grained details. In addition, the Adapter is shown to be essential for transferring the model to downstream medical segmentation tasks.
Overall, these findings validate that DifferSeg is also well suited for binary segmentation scenarios in medical image analysis.

\begin{table}[t]
    \centering 
    \caption{Quantitative Comparison of DifferSeg and the State-of-the-Art SAMs Method on ORSI SOD.}
    \setlength{\tabcolsep}{1mm} 
\resizebox{\linewidth}{!}{
\begin{tabular}{r|cccc|cccc}
\specialrule{1.5pt}{0pt}{0pt}
\multicolumn{9}{c}{\textbf{ORSI SOD}} \\
\midrule
\multirow{2}{*}{Methods}
& \multicolumn{4}{c|}{EORSSD} 
& \multicolumn{4}{c}{ORSSD}  \\

& $S_m \uparrow$ & $F_m \uparrow$ & $E_m \uparrow$& $M \downarrow$
& $S_m \uparrow$ & $F_m \uparrow$ & $E_m \uparrow$& $M \downarrow$

\\ 
\hline 

ORSI-SAM~\cite{RSSOD1}&  .9435  & .8991 & .9842 & .0041 & .9553 &  .9302 & .9897 &.0059  \\
SAM2-RS~\cite{RSSOD2}&  .9472 &  .9013 & .9865 & .0032 & .9555 &  .9317 & .9911 &.0051  \\
 Ours& .9480 &  .9024 & .9874 & .0031 & .9561 &  .9322 & .9919 &.0050 \\
\hline
\end{tabular}
\label{tab:I}
}

\vspace{0.8em} 
\label{tab:8}
\end{table}

\subsection{Generality and Applications}
In this work, we conducted large-scale experiments on 18 popular binary segmentation tasks to demonstrate DifferSeg’s overall capability in multi-task and multi-modal settings. Considering that remote sensing (RS) is also a prominent application, we included a quantitative comparison of DifferSeg and the state-of-the-art SAM-based method on ORSI-SOD in Tab.~\hyperref[tab:8]{8}. 
These results provide preliminary evidence that DifferSeg can be extended to optical remote sensing binary segmentation. However, we agree that broader RS scenarios, such as SAR-optical fusion and multi-spectral segmentation, require further validation.
Future work will investigate the applicability of DifferSeg to remote sensing (RS) segmentation, emphasizing its broad generalization capabilities and potential for integration. Recent related studies~\cite{RS1,RS2,RS3} will be referenced to contextualize its use in RS tasks, further demonstrating the framework’s potential for practical deployment in remote sensing and other complex scenarios.

\section{Discussion and Future Work}

Although DifferSeg demonstrates robust performance and strong generalization ability across multi-task and multi-modal binary segmentation tasks, there remain several directions for further exploration and optimization.

\textbf{Computational Efficiency.}
While SAM2 exhibits strong generalization capabilities, its associated computational cost remains non-negligible~\cite{Eva_SAMs}. As shown in Tab.~\hyperref[tab:3]{3}, although DifferSeg introduces higher FLOPs than SAM2UNet, its measured FPS remains comparable, with 10.0 FPS for DifferSeg and 10.3 FPS for SAM2UNet. This indicates that the increased theoretical computational complexity does not lead to a proportional decrease in practical inference speed. Nevertheless, DifferSeg is not specifically optimized for latency-sensitive edge deployment, and its computational cost may still limit its direct application in real-time scenarios. Future work will focus on further optimizing DifferSeg through model compression techniques, such as pruning~\cite{SAM2pruning}, quantization~\cite{SAM2quan}, or knowledge distillation~\cite{SAM2kd}, to reduce the computational burden for each modality. Additionally, more efficient multi-modal feature fusion strategies can be introduced to reduce redundant computations, thereby improving overall computational efficiency and scalability while maintaining segmentation performance.

\textbf{Scalability to Multiple Modalities.}
The current DifferSeg framework is mainly designed for the common  RGB + 1 Auxiliary Modality setting, such as RGB-D, RGB-T, and RGB-F binary segmentation. This design allows us to validate the effectiveness of complementarity-driven fusion under widely used bimodal segmentation scenarios. However, directly extending the current dual-stream design to three or more modalities may increase the complexity of feature extraction and cross-modal fusion, especially when pairwise interactions among all modalities are exhaustively modeled. Therefore, extending DifferSeg to N-modality perception requires more scalable fusion strategies, such as modality-wise aggregation, shared modality encoders, hierarchical fusion, or token-based unified fusion mechanisms. We regard this as an important direction for future work.

\begin{figure}[!t]
    \centering
    \includegraphics[width=\columnwidth]{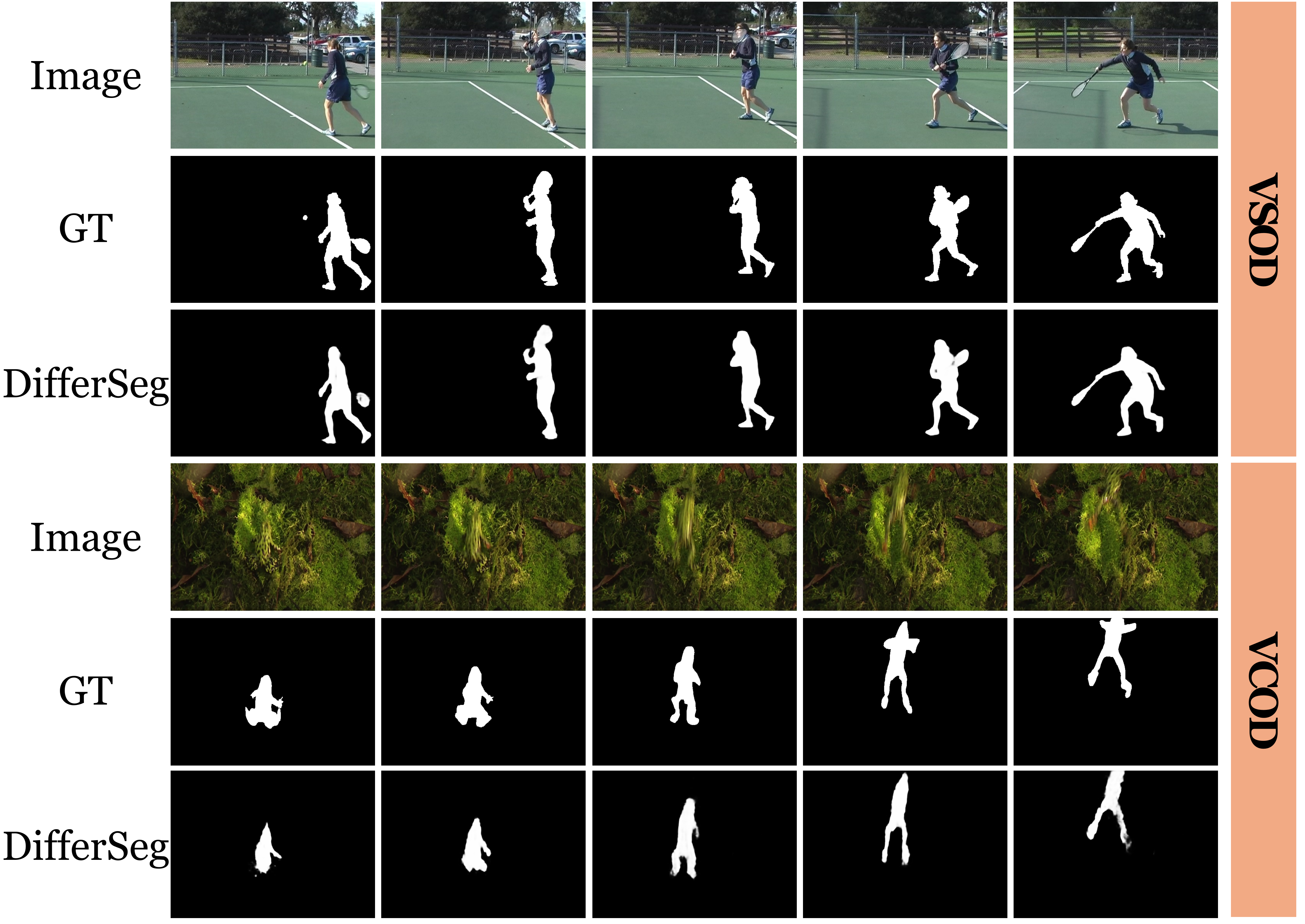}
    \caption{Qualitative video results of DifferSeg. Although DifferSeg does not explicitly model temporal consistency, the visualizations on consecutive frames provide an intuitive assessment of its segmentation behavior in video-related tasks. (GT: ground truth.)}
    
    \label{fig:image9}
\end{figure}
\textbf{Temporal Consistency.}
In video-related tasks, DifferSeg currently treats each video frame as an independent image and does not explicitly model temporal consistency. We acknowledge that this may limit the full potential of SAM2 in video scenarios and could lead to less smooth frame-to-frame predictions in challenging cases. Nevertheless, as shown in Fig.~\hyperref[fig:image9]{9}, 
DifferSeg can still produce reasonable and visually coherent segmentation results on consecutive video frames, demonstrating its applicability to video-related binary segmentation tasks under the current unified framework. Future work could further incorporate temporal information, memory mechanisms, or cross-frame feature modeling strategies to enhance inter-frame consistency and stability, thereby better leveraging the benefits of multi-modal fusion in video scenarios.

\section{Conclusion}
In this paper, we propose DifferSeg, a simple yet general multimodal binary segmentation framework applicable to 18 downstream tasks across both natural and medical image domains. With the introduction of the differential perception fusion module, DifferSeg can adaptively handle modality discrepancies and enhance cross-modal complementarity. Additionally, the frequency-guided decoder effectively balances high- and low-frequency representations, improving the robustness and precision of the decoding process. Extensive experiments demonstrate that DifferSeg generalizes well to diverse binary segmentation scenarios and consistently surpasses state-of-the-art methods across all 18 tasks. We hope that DifferSeg provides a versatile baseline for future multimodal segmentation research and inspires the development of more adaptive and generalizable frameworks. Furthermore, its design principles can guide the community in building unified models that balance modality fusion and feature frequency awareness for a wide range of applications.

\section*{Acknowledgments}
The authors declare that there are no potential conflicts of interest regarding the publication of this paper.
This work was supported by the National Natural Science Foundation of China under Grant Nos. 62262030, 62562039 and 62377011, the Jiangxi Provincial Natural Science Foundation under Grant No. 20232BAB202021, and the Postgraduate Innovation Fund of Jiangxi Normal University under Grant No. YJS202505.

\bigskip
\bibliography{TCSVT.bib}

\begin{IEEEbiography}[{\includegraphics[width=1in,height=1.25in]{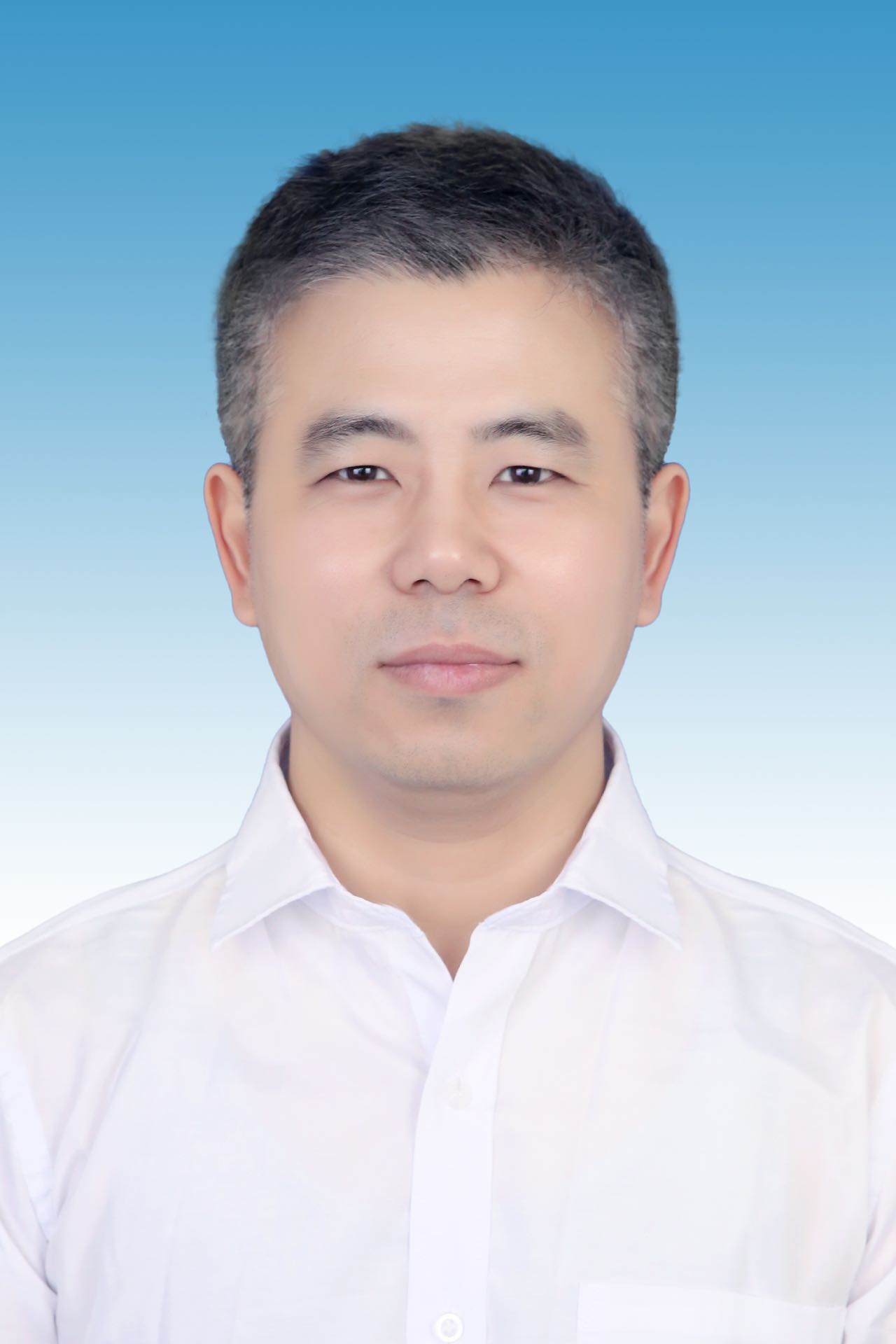}}]
{Qiangqiang Zhou}
received the B.S. degree from Jiangxi Normal University,
Nanchang, China, in 2003, and the M.S. degree in communication and information systems
from Central South University, Changsha, China, in 2009. He received the Ph.D.
degree from Tongji University, Shanghai, China, in 2018. He is currently an Associate Professor with the School of Artificial Intelligence, Jiangxi Normal University, Nanchang, China. His research interests include pattern recognition, machine learning, and computer vision.
\end{IEEEbiography}
\begin{IEEEbiography}[{\includegraphics[width=1in,height=1.25in]{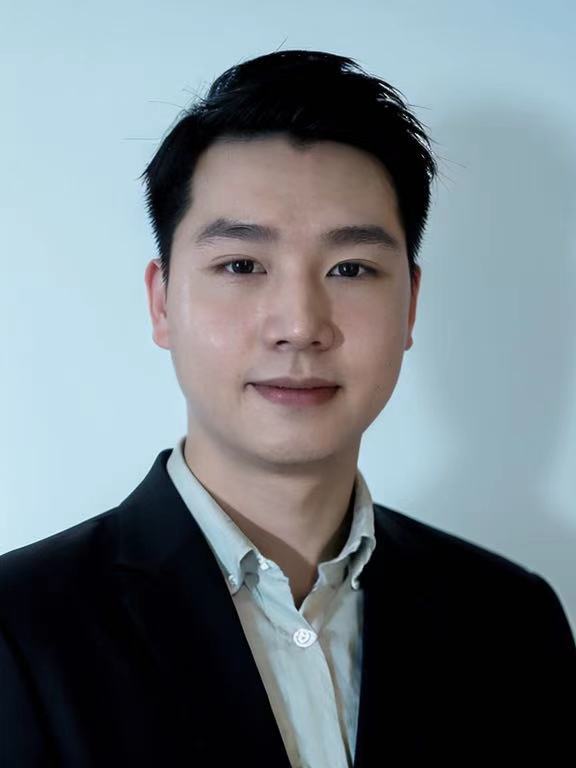}}]
{Jiawei Xu} is a master's student at the School of Artificial Intelligence, Jiangxi Normal University. His research interests include unified context-dependent concept understanding and segmentation, pattern recognition, computer vision, with a specific focus on salient object detection and medical image segmentation.
\end{IEEEbiography}
\begin{IEEEbiography}[{\includegraphics[width=1in,height=1.25in]{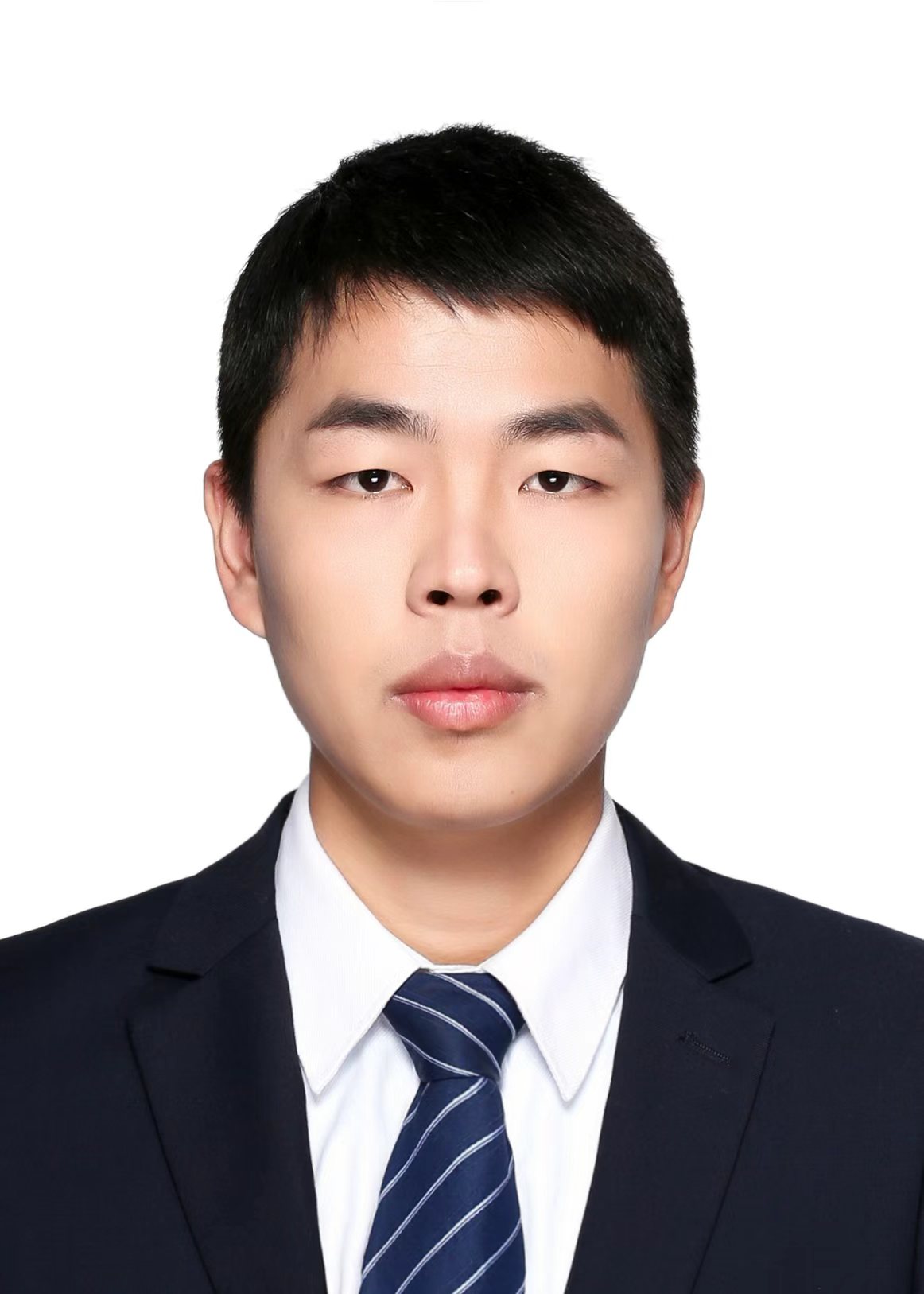}}]
{Yong Chen} received the B.S. degree in School of Science from East China University of Technology, Nanchang, China, in 2015, and the Ph.D. degree from the School of Mathematical Sciences, University of Electronic Science and Technology of China (UESTC), Chengdu, China, in 2020. 
He is currently working with the School of Artificial Intelligence, Jiangxi Normal University, Nanchang, China. From 2018 to 2019, he was a research intern with the Geoinformatics unit, RIKEN Center for Advanced Intelligence Project, Japan. His research interests include hyperspectral image processing, low-rank matrix/tensor representation, and model-driven deep learning. More information can be found on his homepage at: https://chenyong1993.github.io/yongchen.github.io/detection.
\end{IEEEbiography}

\begin{IEEEbiography}[{\includegraphics[width=1in,height=1.25in]{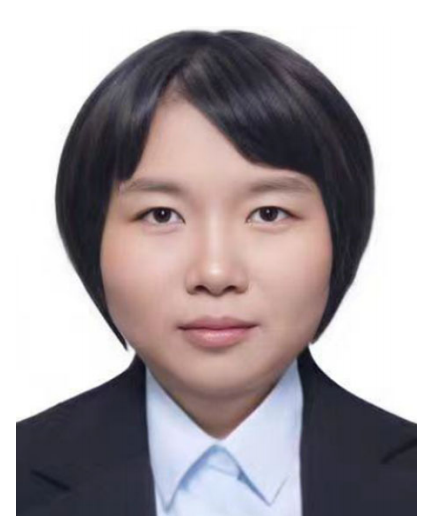}}]
{DandanZhu} received the Ph.D. degree from Tongji University, Shanghai, China, in 2019. She was a Postdoctoral Researcher with the MoE Key Lab of Artificial Intelligence, Shanghai Jiao Tong University, Shanghai, from 2019 to 2021. She is currently a associate professor with the Institute of AI Education, Shanghai, East China Normal University. Her research interests include multi-media signal processing, computer vision and visual attention modeling.
\end{IEEEbiography}

\begin{IEEEbiography}[{\includegraphics[width=1in,height=1.25in]{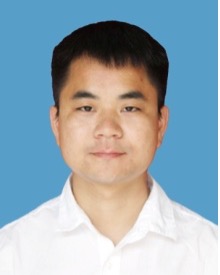}}]
{Yugen Yi} received the Ph.D. degree from the School of Mathematics and Statistics, Northeast Normal University, in 2016. He is currently an Associate Professor with the School of Artificial Intelligence, Jiangxi Normal University. His research interests include artificial intelligence, computer vision, and machine learning.
\end{IEEEbiography}

\begin{IEEEbiography}[{\includegraphics[width=1in,height=1.25in]{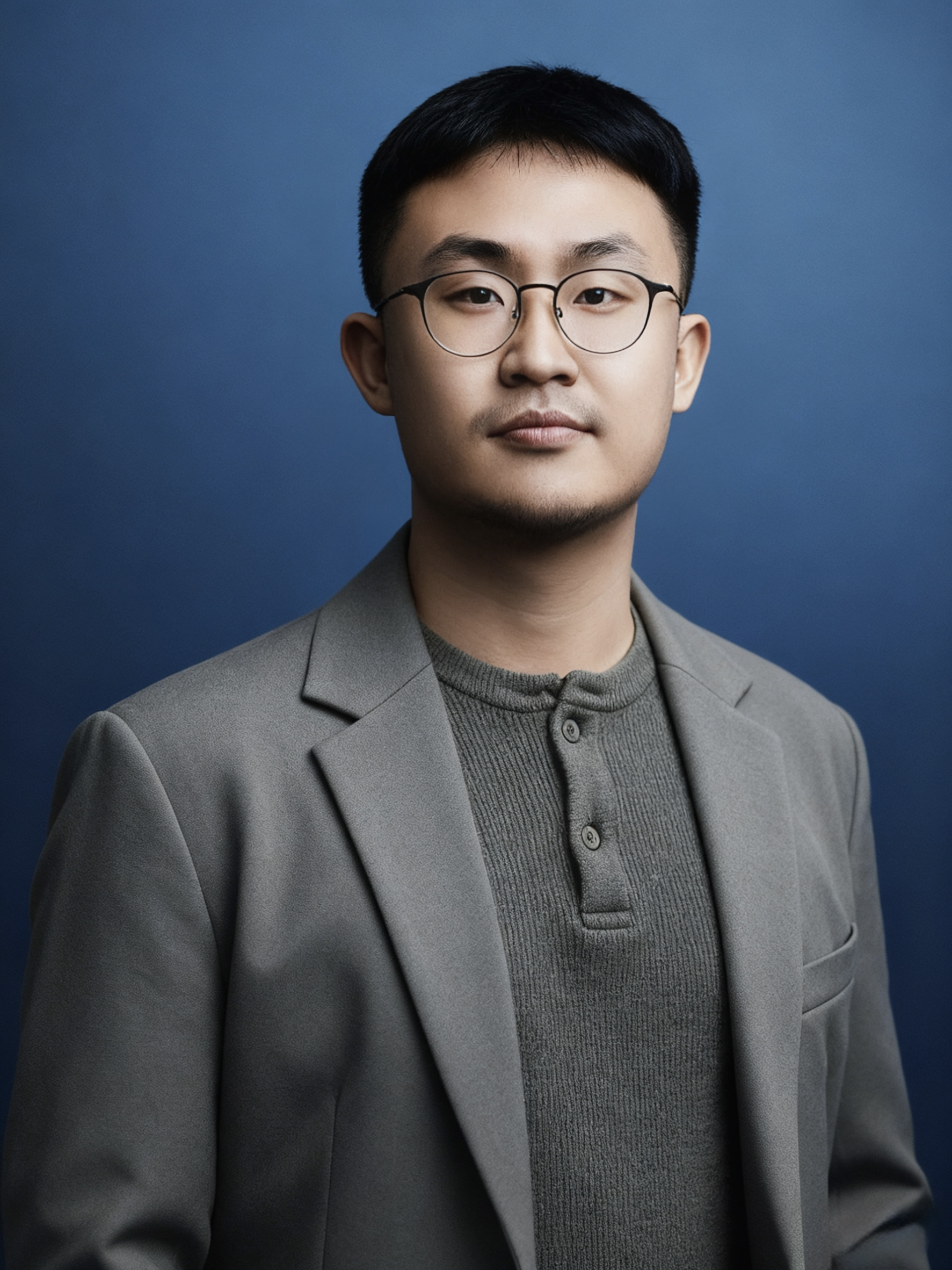}}]
{Xiaoqi Zhao} received the Ph.D. degree from Dalian University of Technology in 2024. He joined Yale University as a postdoctoral researcher in 2025. His research interests include unified context-dependent concept understanding and segmentation, industrial X-ray/CT machine vision (Ai4Industry), medical image analysis (Ai4Health), and self-driven learning mode. He was nominated for the 2025 World Artificial Intelligence Conference (WAIC)  Yunfan Award, recognized as an outstanding reviewer at CVPR 2022, and has presented oral/highlight papers at CVPR, ECCV, and ACM MM.
\end{IEEEbiography}
\end{document}